\documentclass{article}
\usepackage{arxiv}
\usepackage[utf8]{inputenc} % allow utf-8 input
\usepackage[T1]{fontenc}    % use 8-bit T1 fonts
\usepackage{hyperref}       % hyperlinks
\usepackage{url}            % simple URL typesetting
\usepackage{booktabs}       % professional-quality tables
\usepackage{amsfonts}       % blackboard math symbols
\usepackage{nicefrac}       % compact symbols for 1/2, etc.
\usepackage{microtype}      % microtypography
\usepackage{lipsum}		% Can be removed after putting your text content
\usepackage{graphicx}
\usepackage{natbib}
\usepackage{doi}

\usepackage{amsmath} 
\usepackage{amsthm}
\usepackage{amssymb}
\usepackage{nicefrac}

\theoremstyle{definition}
\newtheorem{example}{Example}[section]
\newtheorem{proposition}{Proposition}[section]
\newtheorem{lemma}{Lemma}[section]
\newtheorem{corollary}{Corollary}[section]
\newtheorem{theorem}{Theorem}[section]

\newcommand\Norm[2]{\lVert{#1}\rVert_{#2}}

\newcommand{\Nat}{\mathbb{N}}
\newcommand{\Real}{\mathbb{R}}

\newcommand{\Defeq}{\overset{\mathrm{def}}{=\joinrel=}}
\newcommand{\Def}{\Defeq}

\newcommand{\M}{\mathcal{M}}
\newcommand{\N}{\mathcal{N}}
\newcommand{\Compl}[1]{\sim{#1}}

\newcommand{\Restrict}[2]{{#1}\!\!\restriction_{#2}}

\newcommand{\Value}{\mathit{val}}
\newcommand{\Val}[1]{\Value({#1})}
\newcommand{\Valtilde}[1]{\widetilde{\Value}({#1})}
\newcommand{\Sh}[1]{\mathit{Sh}_{#1}}

\newcommand{\Inner}[3]{\langle {#1} , {#2}\rangle_{#3}}

\newcommand{\PROB}{\mathbb{P}}
\newcommand{\Prob}[1]{{\PROB}({#1})}
\newcommand{\CP}[2]{\Prob{{#1}\,|\,{#2}}}
\newcommand{\E}[1]{\mathbb{E}({#1})}
\newcommand{\CE}[2]{\E{{#1}\,|\,{#2}}}
\newcommand{\Var}[1]{\mathbb{V}({#1})}
\newcommand{\V}[1]{\Var{#1}}
\newcommand{\CV}[2]{\Var{{#1}\,|\,{#2}}}
\newcommand{\Cov}[2]{\mathit{Cov}({#1},{#2})}
\newcommand{\Sobol}[1]{\mathit{S}_{#1}}
\newcommand{\SobolT}[1]{\mathit{T}_{#1}}

\newcommand{\Sample}[2]{{#1}^{({}#2)}}

\newcommand{\Le}[1]{\mathit{L}_{#1}}

\begin{document}

\title{Fairness Analysis with Shapley-Owen Effects}

\date{Aug 28, 2024}	

\author{
\href{https://orcid.org/0000-0002-1405-2990}
Harald Ruess\thanks{The author thanks Chalmers University of Technology and its staff at the department for Computer Science and Engineering for the hospitality and support in preparing these notes during a research visit in August 2024.
} \\
Entalus Computer Science Labs\\
34228-3202 Longboat Key, FL, USA\\
email: \texttt{harald.ruess@entalus.com} \\
url: \texttt{www.entalus.com}
}

% Uncomment to remove the date
%\date{}

% Uncomment to override  the `A preprint' in the header
%\renewcommand{\headeright}{Technical Report}
%\renewcommand{\undertitle}{Technical Report}
\renewcommand{\shorttitle}{Fairness with Shapley-Owen Effects}

%%% Add PDF metadata to help others organize their library
%%% Once the PDF is generated, you can check the metadata with
%%% $ pdfinfo template.pdf
\hypersetup{
pdftitle={Fairness Analysis with Shapley-Owen Effects},
% pdfsubject={q-bio.NC, q-bio.QM},
pdfauthor={Harald Ruess},
pdfkeywords={Fairness in Artificial Intelligence, Uncertainty Quantification, Cooperation Games, Shapley Attribution},
}

%\author{Chih-Hong Cheng}
%\orcid{0000-0002-7265-8413}
%\affiliation{%
%  \institution{Chalmers University of Technology, 
%  Gothenborg, Sweden, \tt{chihhong@chalmers.se}}
%  \department{Computer Science and Engineering}
%  \city{Gothenborg}
%  \country{Sweden}
%}
%\email{chihhong@chalmers.se}

% \date{(Some more or less random notes, Chalmers, Aug 2024)}

% \setcopyright{cc} % CC-BY 4.0

\maketitle

\begin{abstract}
We argue that \emph{relative importance} and its equitable attribution in terms of Shapley-Owen effects is an appropriate one, and, if we accept a small number of reasonable imperatives for equitable attribution, the only way to measure fairness.
On the other hand, the computation of Shapley-Owen effects can be very demanding.
Our main technical result is a spectral decomposition of the Shapley-Owen effects, which decomposes the computation of these indices into a model-specific and a model-independent part. 
The model-independent part is precomputed once and for all, 
and the model-specific computation of Shapley-Owen effects is expressed analytically in terms of the coefficients of the model's \emph{polynomial chaos expansion} (PCE), which can now be reused to compute different Shapley-Owen effects. 
We also propose an algorithm for computing precise and sparse truncations of the PCE of the model and the spectral decomposition of the Shapley-Owen effects, together with upper bounds on the accumulated approximation errors.
The approximations of both the PCE and the Shapley-Owen effects converge to their true values.
\end{abstract}


\vspace{1cm}
\begin{quote}
\hspace*{\fill}\emph{"There is something unfair about a concept of "fairness" that requires a supercomputer $\ldots$"}\\ 
\hspace*{\fill} Deng and Papadimitriou,~\cite{deng1994complexity}
\end{quote}

\section{Introduction}

Machines are widely used to make automated decisions about people, and it is important that these decisions be \emph{fair}, 
that is, without prejudice or favoritism toward any individual or group based on their inherent or acquired characteristics~\cite{cheng2021socially}\@.
Suppose a bank lends money to individuals based on certain characteristics, such as credit score, age group, zip code, and so on~\cite{liu2018delayed}\@. 
The bank wants to maximize its profit by lending money only to those who will repay the loan on time. % - the \emph{true individuals}\@. 
There are sensitive characteristics such as ethnicity or gender that divide the population into groups. 
The bank is considered fair (in lending money) if its 
lending policy is largely agnostic with respect to 
sensitive characteristics.
% In the context of decision-making, an unfair algorithm is one whose decisions are skewed toward a particular group of people~\cite{mehrabi2021survey}\@.
A canonical example of \emph{unfair} algorithmic decision making comes from a software tool 
% used by courts in the United States 
to make pretrial detention and release decisions
that has been shown to be highly 
racially biased~\cite{angwin2022machine}\@.
Similar real-world examples of algorithmic unfairness, along with an analysis of the root causes, have been analyzed in the context of machine learning~\cite{mehrabi2021survey}, since existing bias in real-world data is amplified by machine-learned algorithms\@.

However, a decision can appear fair if it is judged on the basis of individual characteristics without considering their interactions~\cite{buolamwini2018gender}\@.
Crenshaw~\cite{crenshaw2013demarginalizing}, for example, 
analyzes the case of a discrimination suit against black women that was dismissed because the plaintiff had hired women before the Civil Rights Act was passed - albeit white women - during the period when no black women were hired.
Crenshaw~\cite{crenshaw2013demarginalizing} concludes
that a Black woman’s experience of discrimination differs from that of both women and Black people in general~\cite{crenshaw2013demarginalizing}, 
and the interaction along multiple dimensions of identity produces unique and different levels of discrimination for different subgroups.
% \emph{Intersectional fairness} therefore postulates that 
The sum of human experiences of discrimination, in line with Simpson's paradox~\cite{blyth1972simpson},
must take into account the interactions between different characteristics and cannot be limited to considerations of fairness along individual characteristics alone~\cite{gohar2023survey}\@.

% The paradox occurred because women tended to apply to departments with lower admission rates for both genders.
% In general, Simpson's paradox occurs when an association observed in aggregated data disappears or reverses when the same data is disaggregated into its underlying subgroups.
% Similar paradoxon have been observed in a variety of fields, including biology~\cite{chuang2009simpson}, psychology~\cite{kievit2013simpson}, 
% astronomy~\cite{minchev2019yule}, 
% and computational social science~\cite{lerman}\@.

% The \emph{design of experiments} is the design of any task that aims to describe and explain the variation of information under conditions that are hypothesized to reflect the variation. 
% We are exploring the commonalities and differences between fairness and global sensitivity.  

% The decision function itself is assumed to be 
% computable, but is otherwise considered a black box.
% Algorithmic decision making under uncertainty is modeled by random input and response variables for the decision function.
% These considerations motivate the underlying decision model of a random variable $Y = \M(X)$ for a finite sequence $X$ of random input variables.
We are interested in measuring the "fairness" of the response of algorithmic decision making
with respect to certain sensitive subsets of inputs\@.
Fairness constraints are specified by (1) identifying sensitive subsets of input attributes, and (2) determining bounds on the \emph{relative importance} of these subsets to the decision.
Of course, any such specification of \emph{fairness} depends heavily on what is considered to be \emph{acceptable} in a given social context.

Our proposal is a measure of fairness in terms of the \emph{relative importance} of some inputs to the judgment of the decision function.
For example, if \emph{gender} is a sensible attribute of negligible relative importance in a credit decision, then the bank's decision could be said to be fair with respect to gender. 
Similarly, fairness constraints are provided for sensitive subsets such as $\{\mathit{gender}, \mathit{ethnicity}\}$\@.
We postulate that the variance of the expected output, conditioned on the sensitive inputs, is a natural measure of \emph{relative importance}\@. 
% More formally, for a given subset $u$ of input indices we define the relative importance of inputs $(X_i)_{i \in u}$ as
%   $
%       \Var{\CE{\M(X)}{X_u}}
%   $\@.
In  this way, we can also specify a larger class of fairness constraints in terms of ratios and differences in the relative importance of different sensitive inputs,
including~\emph{disparate impact}~\cite{feldman2015certifying}
and~\emph{differential fairness}~\cite{dwork2012fairness}\@.
The US Equal Employment Opportunity Commission, for example, advocates
% a $80\%$ limit on the \emph{disparate impact} of sensitive 
% attributes in hiring decisions\@.
\emph{disparate impact} in the sense
that the ratio of the chances of being hired if the sensitive trait is met to the chances of being hired if the sensitive trait is not met shall be bounded from above by $0.8$\@.\footnote{US Equal Employment Opportunity Commission\@.\emph{Uniform guidelines on employee selection procedures}, March 2, 1979.}

Relative importance, when normalized by the variance of the decision function, is consistent with the notion 
of a \emph{Sobol' index} for subsets of inputs\@.
In~\emph{global sensitivity analysis},
 these indices are useful for quantifying which 
inputs most influence response variability.  
However, Sobol' indices are difficult to interpret in the presence of statistical dependence between inputs~\cite{iooss2019shapley}\@. 

We are thus interested in an \emph{importance attribution} of the relative importance of input characteristics such that
(1) the variance of the decision function is distributed among, and thus explained by, the attributed relative importance of all of the inputs, 
(2) inputs of equal relative importance receive equal attribution,
(3)  inputs with a zero marginal contribution to the relative importance of all supersets receive a zero attribution,
and (4)
each pair of inputs should share equally in the gain in relative importance for all possible shared interactions.
Since \emph{Shapley values}~\cite{shapley1953value,winter2002shapley}
are the only attribution that satisfies these constraints, 
the scenario is as follows. 
We set up a \emph{cooperation game}~\cite{osborne1994course,peleg2007introduction,chalkiadakis2022computational} with the inputs as the players of the game and 
subsets of the inputs as possible \emph{coalitions} of players\@.
The value of each coalition is measured by its relative importance to the variation in expected output.
Now, the Shapley value assigns the mutual contribution -- due to correlation and interaction -- of coalitions to each individual input within the coalition in a manner that is consistent with the above imperatives (1)-(4)\@.

Shapley values for relative importance games are also called Shapley effects~\cite{owen2014sobol,song2016shapley}\@.
% Shapley effects are bracketed by Sobol indices, assuming independence, as a lower bound and by the total Sobol' indices as an upper bound.
% Therefore, whenever their difference is small enough, the Shapley effect can be approximated by Sobol' indices.
A generalization of Shapley effects to the
Shapley-Owen~\cite{owen1972multilinear,owen2014sobol} interaction effects attributes a fair share of relative importance not only to a single input but, more generally, to all possible subsets of the inputs.
Using this index, it is now also possible to gain insight into the \emph{synergistic} or \emph{antagonistic} nature of interactions within subsets of intputs. 
For example, the Shapley-Owen effect for $\{\mathit{Gender}, \mathit{Ethnicity}\}$ attributes its marginal contribution to the relative importance of all its supersets.

In practice, we may not be able to determine the exact Shapley and Shapley-Owen interaction indices because there is an exponential number of possible coalitions of inputs, and computing Shapley values is already hard for some simple games~\cite{deng1994complexity,conitzer2006complexity}\@.
However, several algorithms have been proposed to approximate Shapley values~\cite{mann1962evaluating,owen1972multilinear,bachrach2010approximating,fatima2008linear,maleki2013bounding} by considering only a 
sample of coalition values, thus avoiding the need to consider an exponential number of such values.
Shapley effects can also be estimated~\cite{song2016shapley,broto2020variance}\@,
and Shapley effects are easily computable in 
%the Gaussian linear framework
some special cases~\cite{owen2017shapley,iooss2019shapley,broto2019sensitivity,broto2020variance,broto2020gaussianlinearapproximationestimation}\@.

Here we follow a different path in that we are separating 
the computation of Shapley-Owen effects into model-independent and model-dependent computations. 
Hereby, the model-independent computations are considered to be pre-computed, once-and-forall, and the model-dependent computation reduces to the construction of a sufficiently precise \emph{polynomial chaos expansion} (PCE)\@,
which relies on the  \emph{spectral decomposition} of the decision function\@. 
In other words, Shapley-Owen effects are expressed analytically in terms of the PCE coefficients of the underlying decision model.
This PCE can be reused for different sensitivity analyses as long as the underlying decision function and the input distribution are unchanged. 
% Algorithms for constructing precise and
% sparse PCEs are developed in the emerging field of 
%\emph{uncertainty quantification}
% ~\cite{sullivan2015introduction}\@.

This paper is structured as follows. 
Section~\ref{sec:prelims} reviews some basic mathematical notation and concepts.
Section~\ref{sec:importance} motivates and formalizes the central notion of \emph{relative importance} as the basis for specifying fairness constraints. We also argue that the related notion of Sobol' sensitivity indices are not sufficient for fairness analysis in the important case of dependent inputs. 
Instead, Section~\ref{sec:soe.and.fairness} proposes Shapley-Owen effects, because of their "fair" attribution of relative importance, as adequate measures for specifying fairness constraints.
In order to make this paper as self-contained as possible we review basic concepts of PCE in~Section~\ref{sec:pce}\@. 
Based on these development we are proposing a simple algorithm for constructing precise and sparse truncations of PCE with known truncation errors. 
Section~\ref{sec:computing.soe} contains our main result, which is a spectral decomposition of Shapley-Owen effects based on the PCE of the underlying decision function.
We put these developments into perspective and compare our results with related work in Section~\ref{sec:related}.
Finally, Section~\ref{sec:conclude} concludes with some final remarks.

\section{Preliminaries}\label{sec:prelims}

Our developments are based on standard developments and notations of \emph{probability theory}, \emph{functional analysis}, and the emerging field of \emph{uncertainty quantification}~\cite{xiu2010numerical,sullivan2015introduction}\@.
The \emph{probability space} $(\Omega, \mathcal{E}, \PROB)$
consists of (1) the 
domain $\Omega \Defeq \Omega_1 \times \ldots \times \Omega_d$ of dimension $d \in \Nat$,
(2) the $\sigma$-algebra $\mathcal{E}$ on $\Omega$,
and (3) the probability measure $\Prob{E}$ 
% on $(\Omega, \mathcal{E})$
for all measurable events $E \in \mathcal{E}$\@.
For $B \in \mathcal{E}$ with $\Prob{B} > 0$ the
\emph{conditional probability measure} $\Prob{.\,|\, B}$
on $(\Omega, \mathcal{E})$ is defined by 
$\Prob{E\,|\, B} \Defeq \nicefrac{\Prob{E \cap B}}{\Prob{B}}$\@,
for all events $E \in \mathcal{E}$\@.

A \emph{random variable} in $(\Omega, \mathcal{E}, \PROB)$  is a function $X: \Omega \to \Real$,
which is \emph{measurable} in the sense that 
$X^{-1}(B) \in \mathcal{E}$ for all events $B$ in the Borel-$\sigma$-algebra on the reals\@.
The \emph{probability density function} (pdf) $f: \Omega \to \Real$ for the random variable $X$ is defined as $f(x) \Def \Prob{X = x} \Def \Prob{\{\omega \in \Omega \,|\, X(\omega) = x\}}$\@.
We assume that the pdf $f$ for $X$ is measurable.
% The pdf $f(.)$ is assumed to be \emph{measurable} in the sense that
% for all $B \in \mathcal{B}$ the pre-images $f^{-1}(B)$ are contained in the 
% $\sigma$-algebra $\mathcal{E}$\@.
The \emph{marginal probability measures} $\PROB_i(.)$, for $i = 1, \ldots, d$,
are defined as $\PROB_i(E_i) \Defeq \Prob{\Omega_1 \times \ldots \times E_i \times \ldots \times \Omega_d}$, which assumes that the underlying projections $\Omega \to \Omega_i$ are measurable\@.
In this case, the \emph{marginal pdfs} $f_i(.)$ are given 
by $f_i(x_i) \Defeq \PROB_i(X_i = x_i)$ for $x_i \in \Omega_i$\@.
The random variables $X_i$ are \emph{independent} 
if and only if $f(x) = \Pi_{i=1}^d f_i(x_i)$, for $x = (x_1, \ldots, x_d) \in \Omega$\@.
% if and only if $\PROB = \PROB_1 \otimes \ldots \otimes \PROB_d$\@.
%
For measurable $\M, \N: \Omega \to \Real$
the \emph{inner product} weighted by the measurable pdf $f$ is defined as
  \begin{align}\label{def:inner}
  \Inner{\M}{\N}{f}
    &\Defeq \int_{\Omega} \M(x)  \N(x)  f(x) \,dx\mbox{\@,}
  \end{align}
and $\M$ and $\N$ are said to be \emph{orthogonal} if 
$\Inner{\M}{\N}{f} = 0$\@.
The inner product~(\ref{def:inner}) induces the \emph{norm}
  % \begin{align}\label{norm}
     $\Norm{\M}{f} \Defeq \sqrt{\Inner{\M}{\M}{f}}$\@.
 %   \end{align}
% and a random variable $X: \Omega \to \Real$ with pdf $f$ is 
% \emph{square-integrable} if $\Norm{X}{f} < \infty$\@.
The \emph{Lebesgue space} $L^2(\Omega, \PROB; \Real)$ (short: $L^2(\Omega)$)
is the set of \emph{random variables} $X: \Omega \to \Real$
that are \emph{square-integrable} such that $\Norm{X}{f} < \infty$\@.
This space is complete with respect to the induced norm, so $L^2(\Omega)$ is a \emph{Hilbert space}\@.
For a random variable $X$ with measurable pdf $f$, 
the expectation $\E{X}$ is the inner product $\Inner{X}{\mathbf{1}}{f}$,
where $1$ is a random variable with pdf $f_{\mathbf{1}}(1) = 1$,
and $\Var{X} \Defeq \Norm{X}{f}^2$ is the \emph{variance} of $X$\@. The random variables for the conditional expectation $\CE{X}{Y}$ and the conditional variance $\CV{X}{Y}$ are also defined as usual.

A \emph{multi-index} is an element $\alpha \in \Nat^d$,
and we write $0$ for the multi-index $(0, \ldots, 0)$\@. 
%A partial order on multi-indices is defined by
% $\alpha \leq \beta$ if and only if $\alpha_i \leq % \beta_i$ for all $i = 1, \ldots, d$\@.
For a given multi-index $\alpha = (\alpha_1,\ldots, \alpha_d)$, 
a \emph{monomial} $X^\alpha$ is of the form $X_1^{\alpha_1} \ldots X_d^{\alpha_d}$\@.
A polynomial in $\Real[X]$
is a finite linear combination of monomials $X^\alpha$\@.
The \emph{total degree} $|\alpha|$ of a multi-index $\alpha$ is the sum $\sum_{i=1}^d \alpha_i$\@.
For the random variable 
$X = (X_1, \ldots, X_d)$ in
$(\Omega, \mathcal{E}, \PROB)$
we are also interested in subsequences $X_u \Def (X_i)_{i \in u}$, where $u \subseteq [1,d]$\@. 
Depending on the context of use, $X_u$ is also interpreted as a set.
In case, $i \notin u$ we also use the shorthand  $u + i$
for $u \cup \{i\}$,
and $\Compl{u}$ denotes the set difference $[1,d] \setminus u$\@. 
For a polynomial $P(x) \in \Real[x]$, the \emph{projection} function
\begin{align}\label{def:characteristic.function}
 \pi_u(P(x)) &\Def 
     \begin{cases}
         P(x) &\text{if~} X_u = \mathit{vars}(P(x)) \\
         0 &\text{otherwise}\mbox{\@.}
     \end{cases}
\end{align}
retains only the polynomials in which exactly 
the subset $X_u$ of variables occur.
The \emph{characteristic} function $\chi_u(P(x))$ is
defined similarly, but it returns $1$ 
instead of the argument $P(x)$\@.

\section{Relative Importance}\label{sec:importance}

Algorithmic decisions under uncertainty are modeled by a discrete or real-valued random variable $Y = \M(X)$ % \in L^2(\Omega)$ 
for  $X \Def (X_1, \ldots, X_d) \in L^2(\Omega)$ a finite sequence of random input variables $X_i$ for $i = 1, \ldots, d$
with domain $\Omega = \Omega_1 \times \ldots \times \Omega_d$
and measurable \emph{pdf} $f(.)$\@.
The variance $\sigma^2 \Def \Var{\M(X)}$ of $\M(X)$ is finite.
Moreover, the decision function $\M$ is assumed to be \emph{computable}, but is otherwise considered a black box.
% for $x = (x_1, \ldots, x_d)$ and $x_i \in \Omega_i$,
% and $\M(X) \in L^2(\Omega)$\@. 

Now, given any subset $u \subseteq [1,d]$, the 
\emph{relative importance} (alternatively, \emph{value} or \emph{explanatory power}) $\Val{u}$ of the inputs in $X_u$ for the outcome $\M(X)$ is conveniently measured by the variance of the expected outcome conditioned on $X_u$\@:
 \begin{align}\label{def:value}
   \Val{u} &\Def \V{\CE{\M(X)}{X_u}}\mbox{\@.}
   \end{align}
Here, the empty set creates no value and the entire set of inputs contributes the variance $\sigma^2$ of $\M(X)$\@.
By the \emph{total law of variance}, 
$\sigma^2 =  \Val{u} + \Valtilde{u}$\@, for
\begin{align}
   \label{def:valuetilde}
   \Valtilde{u} &\Def \E{\CV{\M(X)}{X_u}}\mbox{\@.}
   \end{align}
There is a close relationship between relative importance and \emph{Sobol' global sensitivity indexes}\@.
In the case of independent inputs $X$, at least, 
$\M(X) = \sum_{u \subseteq [1,d]} \M_u(X_u)$
the \emph{Hoeffding decomposition}~\cite{hoeffding1948} states that there are pairwise orthogonal $\M_u(X_u)$ with (Appendix~\ref{app:anova})
\begin{align}\label{hoeffding}
    \M(X) &= \sum_{u \subseteq [1,d]} \M_u(X_u)\mbox{\@.} 
\end{align}
This decomposition of $\M(X)$ is unique,
% \begin{align}\label{cond.exp}
    $\CE{\M_u(X_u)}{X_i} = 0$\@,
 %   \label{orthogonal}
and  $\E{\M_u(X_u)\M_v(X_v)} = 0$\@,
% \end{align} 
for all $u,v \subseteq [1,d]$ and $i \in u$\@.
%Using the identity~(\ref{cond.exp}) the terms of the Hoeffding decomposition can be computed recursively as
%  \begin{align}\label{eq:recursive.terms}
%      \M_u(X_u) &= \CE{\M(X)}{X_u} - \sum_{v \,:\,v \subsetneq u}
%                        \CE{\M(X)}{X_v}\mbox{\@,}
%  \end{align}
%In particular, $\M_i(X_i) = \CE{\M(X)}{X_i}$\@.
Therefore, the expected value of $\M(X)$ is given by 
$\M_{\emptyset}(X_{\emptyset}) = \E{\M(X)}$\@.
The central property of the \emph{analysis of variance} (ANOVA) is the decomposition 
\begin{align}\label{ANOVA}
    \sigma^2 &= \sum_{u:\, \emptyset \neq u \subseteq [1,d]}\sigma_u^2
\end{align}
of the total variance of $\M(X)$,
where $\sigma_u^2 \Def \V{\M_u(X_u)}$ denotes 
the variance of the partial effect $\M_u(X_u)$\@.
The ANOVA decomposition~(\ref{ANOVA}) is a direct consequence of the orthogonality of the partial effects $\M_u(X_u)$ in the Hoeffding decomposition\@.
The best predictor of $\M(X)$ given $X_u$ 
is, by definition, the conditional expectation
  %  \begin{align}\label{best.predictor}
    $\CE{\M(X)}{X_u} = \sum_{v:\, v \subseteq u} \M_v(X_v)$\@.
  %  \end{align}
The \emph{Sobol' indices} $\Sobol{u}$ 
% represents the variance explained by $X_u$ of the expected value of $\M(X)$\@.
are obtained by taking the variance on both sides of the identity for the best predictor. % identity~(\ref{best.predictor}).\footnote{A normalization by $\sigma^2 ={\Var{\M(X)}}$ is frequently used in the definition of Sobol' indices, but this  normalization is not needed here.}  
    \begin{align}\label{def:sobol}
   \Sobol{u} &\Def \Val{u} = \sum_{v:\,v \subseteq u} \sigma_v^2\mbox{\@.}
   \end{align}
The interpretation of the Sobol’ indices is easy
in the case of independent inputs,
because the variance decomposition~(\ref{ANOVA}) of $\M(X)$ is unique. 
Thus, the first-order Sobol’ index $\Sobol{i}$, for $i \in [1,d]$\@, represents the amount of the output variance due to input $X_i$ alone\@. 
The second-order Sobol’ index $\Sobol{\{i, j\}}$ expresses the
contribution of the \emph{interactions} of the pairs of 
variables $X_i$ and $X_j$, and so on
for the higher orders. As the sum of all Sobol’ indices is equal to the variance $\sigma^2$ of $\M(X)$, the indices are interpreted as proportions of explained variance.
When $\Sobol{u}$ is large, it means that the combined effect of all $X_j$ for $j \in u$ makes an important contribution to the variance of the expected outcome of $\M(X)$\@.
It therefore is the natural choice for measuring the \emph{relative importance} of the inputs $X_u$ on $\M(X)$\@.
% From the M\"obius inversion of~the identity (\ref{def:sobol}) one obtains
% $\sigma^2_u = \sum_{v \subseteq u} (-1)^{|u|-|v|} \V{\CE{\M(X)}{X_v}}$\@.
%The Sobol' index $\Sobol{u}$ can also be rewritten as $\nicefrac{1}{\sigma^2} \cdot \Cov{\M(X)}{\M_u(X_u)}$~\cite{janon2014asymptotic}\@, which leads to the estimator described in~\cite{saltelli2000sensitivity}\@.
%
The \emph{total Sobol'} index 
 \begin{align}
   \label{def:sobolT}
   \SobolT{u} &\Def \Valtilde{\Compl{u}}
               = \sum_{v \,:\,v \cap u \neq\emptyset}\sigma_v^2 % \mbox{\@.}
   \end{align}
expresses the “total” sensitivity of the variance 
$\sigma^2$ of $\M(X)$ to the inputs in $X_u$\@. 
It can be interpreted as the expected remainder of the variance once the variables $X_{\Compl{u}}$ are known.
If $\SobolT{u}$ is small, it means that the joint effects of $X_j$ for $j \in u$ make little difference, even when all interactions between them and $X_k$ for $k \notin u$\@ are taken into account.
% Sometimes this means that these variables are so unimportant that they can be "frozen" at a fixed value. 
% value, thus reducing the dimension of the domain of $\M$\@.
In the case of input independence, the inequality
$\Sobol{u} \leq \SobolT{u}$ between Sobol' indices follows directly from the identities~(\ref{def:sobol}) and~(\ref{def:sobolT})\@.
% \begin{align}
%   \Sobol{u} &\leq \SobolT{u}
% \end{align}
Moreover, 
$\Sobol{u} = \sigma^2 - \SobolT{\Compl{u}}$\@.
% since
% $\Sobol{u} = \V{\CE{\M(X)}{X_u}}$ equals, by the \emph{total law of variance}, 
% $\sigma^2 - \E{\CV{\M(X)}{X_u}}$\@.
% = \sigma^2 - \SobolT{\Compl{u}}$\@.
% \begin{align}\label{Sobol-duality}
%   \Sobol{u} &= \sigma^2 - \SobolT{\Compl{u}}\mbox{\@.}
% \end{align}
% Therefore, a total Sobol' index may also be computed from a corresponding Sobol' index.

% Using equality~(\ref{eq:recursive.terms}) we obtain
% \begin{align}
%   \Sobol{u} &= \Var{\CE{Y}{X_u}}\mbox{\@.}  
%   \end{align}
% \noindent
% For the special case of first-order Sobol indeces $S_i \Def S_{\{i\}}$ and $\SobolT{i} \Def \SobolT{\{i\}}$ one obtains 
% \begin{align}\label{eq:fo-sobol}
%      S_i %&= \V{\CE{\M(X)}{X_i}} 
%          % = \sigma^2 - \E{\CV{\M(X)}{X_i}}  
%           &= \sigma^2_i \\
%     \SobolT{i} % &= \sigma^2 - \V{\CE{\M(X)}.  
%                % {X_{\Compl{i}}}}
%                % = \E{\CV{\M(X)}{X_{\Compl{i}}}}
%       &= \sum_{v \,:\,i\in v} \sigma^2_v  \mbox{\@.} 
%       \label{SobolT}
% \end{align}
% $S_i$ therefore quantifies the additive effect of the input $X_i$ separately\@. 
% Neither $\Sobol{i}$ nor $\SobolT{i}$ sum to $\Sobol{D} = \sigma^2$, and so fail the 
% efficiency property for Shapley values.
% The equality~(\ref{SobolT}) is not practical for actual 
%computations since it would result in computing
%each index separately. 
%But, for equality~(\ref{Sobol-duality}), the total index can
%be rewritten as $\SobolT{i} = \sigma^2 - \Sobol{\Compl{i}}$\@.
%From this, the idea of computing indices of groups of % variables is developed.

% \paragraph{Sobol' indeces for correlated features}
A common way to deal with input dependence is to define a
Hoeffding-like decomposition for dependent inputs and then define variable importance 
through this generalization~\cite{chastaing2012generalized,chastaing2015generalized,idrissi2023hoeffding}\@.
However, the dependent-variable Hoeffding decomposition
in~\cite{chastaing2012generalized,chastaing2015generalized}
imposes strong constraints on the joint probability distribution of the inputs that already fail for  non-zero correlation Gaussians, 
and the resulting importance of a variable can be negative~\cite{chastaing2015generalized}\@.
This is conceptually problematic, since a variable on which the function does not depend at all has zero importance, and is therefore more important than a variable on which the function does depend in a way that gives it negative importance.
Many variants of the Sobol' indices have been proposed for
dependent inputs~\cite{xu2008uncertainty,mara2012variance,chastaing2012generalized,chastaing2015generalized,mara2015non,kucherenko2012estimation},
but Sobol’ indices present a difficult interpretation in the presence of input dependence~\cite{iooss2019shapley}\@. 

%The \emph{analysis of covariance} (ANCOVA) is a way to generalize the formulation of the Sobol' indices for correlated input variables~\cite{xu2008uncertainty}\@. It is based on a
%covariance decomposition method similar to the \emph{analysis of variance} (ANOVA)\@.
%Alternatively, \emph{Kucherenko indices} generalize the Sobol’ 
%indices to the case of dependent input features using a direct
%decomposition of the output variance with the law of total variance\cite{kucherenko2012estimation}\@.

\section{Shapley-Owen Effects and Fairness}\label{sec:soe.and.fairness}

% Shapley Effects are designed to address the conceptual difficulty of measuring the relative importance of dependent variables using ANOVA.
Shapley effects overcome these conceptual problems of Sobol' indices by attributing the mutual contribution - due to correlation and interaction - of a subset of inputs to each individual input within that subset~\cite{owen2014sobol,song2016shapley,owen2017shapley}\@. 
Like ANOVA, they use variances, but unlike ANOVA for dependent data, a Shapley effect never goes negative, and it can be defined without making onerous assumptions about the input distribution.

\subsection{Shapley-Owen Effects}

In economics, Shapley values are often used to solve the attribution problem, where the value created by the collaborative efforts of a team needs to be "fairly" attributed to the individual members of that team.
In our setting, the \emph{team} is the set $\{X_1, \ldots, X_d\}$
of input variables, the value of any 
subset $X_u$ of variables is its \emph{relative importance} $\V{\CE{\M(X)}{X_u}}$ on the variance of outcomes $Y = \M(X)$\@,
and Shapley values are used to attribute the relative importance of subsets of inputs in a "fair"  way, which will be explained in more detail below.

Let $\Val{u} \in \Real$ be the value attained by the subset 
$u \subseteq \{1, \ldots, d\} \equiv [1,d]$\@.
It is always assumed  that $\Val{\emptyset} = 0$.
\newcommand{\UG}{\chi}
The Shapley-Owen value 
% generalizes the notion of Shapley values from individual input importance to any 
takes into account the possible interaction between inputs~\cite{owen1972multilinear,grabisch1999axiomatic}\@.
  \begin{align}\label{def:shapley.owen}
   \Sh{u}(\Value) &\Def \frac{1}{(d - |u| + 1)} 
     \sum_{v \,:\,v \subseteq \Compl{u}} 
        \binom{d - |u|}{|v|}^{-1}\,
  %   \frac{(d - |v| - |u|)!\, |v|!}{(d - |u| + 1)!} \,
                    \sum_{w \,:\,w \subseteq u} (-1)^{|u| - |w|} \Val{v + w}
  \end{align}
By using this index it is therefore possible to gain insights on the synergistic/antagonistic nature of interactions~\cite{rabitti2019,plischke2021computing}\@.
The same Shapley-Owen value arises if we use values
$\Valtilde{\Compl{u}}$~(\ref{def:valuetilde})
% $\E{\CV{\M(X)}{X_{\Compl{u}}}}$ 
instead of the relative importance~(\ref{def:value})~\cite{song2016shapley}\@.
When $u = \{i\}$ the Shapley-Owen value (\ref{def:shapley.owen}) reduces to the Shapley value of a single input~(\ref{def:shapley})\@.
The \emph{Shapley} value $\Sh{i}(\mathit{val})$ for 
$i = 1, \ldots, d$ is
  \begin{align} \label{def:shapley}
      \Sh{i} &= \frac{1}{d}
               \sum_{v \,:\, v \subseteq \Compl{i}} {\binom{d - 1}{|v|}}^{-1} 
               (\Val{v + i} - \Val{v})\mbox{\@.}
  \end{align}
% where $u + i$ is shorthand for $u \cup \{i\}$\@. 
The \emph{marginal contribution} $(\Val{v + i} - \Val{v})$ is how much the model output changes when a new feature $i$ is added,
the \emph{combinatorial weight} ${\binom{d - 1}{|v|}}$$^{-1}$ is the weight given to each of the different subsets of inputs with size $|v|$\@,
and \emph{averaging} by $\nicefrac{1}{d}$ determines the average of all marginal contributions from all possible subsets of sizes ranging from $0$ to $d-1$\@.

If \emph{relative importance}
   %\begin{align*}
   %    \Val{u} &\Def \Sobol{u}\mbox{\@,}
   %\end{align*}
 %  $\Val{u} \Def \Sobol{u}$\@,
%where $\Sobol{u}$ is the Sobol' index for $u$ as defined in~(\ref{def:sobol}),
is used to define the game
then~$\Sh{u}(\Value)$~(\ref{def:shapley.owen}) 
is also said to be a \emph{Shapley-Owen effect} of the subset $X_u$ of inputs\@.
Similarly, the special case $\Sh{i}(\Value)$ is called the \emph{Shapley effect} of the input $X_i$\@.
The computation of both Shapley and Shapley-Owen effects can be very demanding because they require summation over an exponential number of subsets, and computing the relative importance for all these subsets can be time-consuming~\cite{song2016shapley,plischke2021computing}\@.
For some special cases, though, they are relatively easy to calculate (Appendix~\ref{app:so.effects})\@.

\subsection{Fairness}

The following propositional logic example is an illustration of Shapley effects and their interpretation in terms of fairness.
\begin{example}[\cite{huang2023inadequacyshapleyvaluesexplainability}]\label{ex:prop}
Consider a three-player game $D \Def \{1,2,3\}$ of value
$\Val{u} \Def \Sobol{u} = \V{\CE{Y}{X_u}}$\@, where 
   $$
   Y \Def \M(X) \Def \M(X_1, X_2, X_3) = (X_1 \land X_2) \lor (\neg X_1 \land X_3)\mbox{\@,}
   $$
$X_i \in \{0, 1\}$ independent with $\Prob{X_i = x_i} = \nicefrac{1}{2}$ for $i = 1, 2, 3$\@.
  \begin{align*}
    \CE{Y}{X_u = x_u} 
      &= \CP{Y = 1}{X_u = x_u}
      = \frac{\Prob{Y = 1, X_u = x_u}}{\Prob{X_u = x_u}}
      = 2^{|u|}\, \Prob{Y = 1, X_u = x_u}
  \end{align*}
The joint probability 
$\Prob{Y = 1, X_u = x_u}$
is equal to the number of satisfying assignments of $Y = \M(X)$, where $X_u = x_u$ are fixed, divided by the size $2^{3}$ of the input space\@.
For example,
  \begin{align*}
  \CE{Y}{X_1 = 0} 
     = 2\, \Prob{Y = 1, X_1 = 0} 
     = 2^{1 - 3} \,2
     = \nicefrac{1}{2}\mbox{\@,} 
  \end{align*}
since there are exactly $2$ satisfying assignments with $X_1 = 0$\@.
Similarly, $\CE{Y}{X_1 = 1} = \nicefrac{1}{2}$\@.
From the \emph{total law of expectations} it follows that
$\E{\CE{Y}{X_u}} = \E{Y} = \nicefrac{1}{2}$ for all subsets of inputs $X_u$, since $Y$ has $4$ (out of $8$) satisfying assignments\@.\footnote{
The expectation $\E{\CE{Y}{X_u}}$
of the random variable $\CE{Y}{X_u}$
is taken with respect to $X$\@. 
} For example, the variance of the random variable $\CE{Y}{X_1}$ is computed as
  \begin{align*}
 %     \E{\CE{Y}{X_1}} &= \nicefrac{1}{2} \, \CE{Y}{X_1 = 0}
 %                      + \nicefrac{1}{2}  \, \CE{Y}{X_1 = 1} 
 %                     = \nicefrac{1}{2} \\
     \V{\CE{Y}{X_1}} &= \E{(\CE{Y}{X_1} - \nicefrac{1}{2})^2}  
                     = \nicefrac{1}{2} \, (\CE{Y}{X_1 = 1} - \nicefrac{1}{2})^2 + \nicefrac{1}{2} \, (\CE{Y}{X_1 = 0} - \nicefrac{1}{2})^2  
                     = 0
  \end{align*}
% In general (???), $\CE{Y}{X_u = x_u}$
% equals $2^{|u| - |D|}$ times the number of satisfying assignments of $\M(X)$\@, where the subset $X_u$ is fixed to some assignment $x_u$\@.
%  \V{\CE{Y}{X_u = x_u}}
%      &= \sum_{u \in D} 2^{-|u|}\, (\CE{Y}{X_u = % x_u} - \mu)^2  
%      = \sum_{u \in D} 2^{-|u|}\,
%              (2^{|u|}\, \Prob{Y = 1, X_u = x_u} - \mu)^2 \\
%   \mu &\Def \sum_{u \in D} 2^{-|u|} \CE{Y}{X_u = x_u} = 
%\sum_{u \in D} \Prob{Y = 1, X_u = x_u}\@.       
%   \end{align*}
% Notice that $\mu$ is counting the satisfying assignments of $\M(X)$\@.
% Now, $\CE{Y}{X} = 1$ with probability $\nicefrac{1}{2}$, 
%since 4 out of 8 possible assignments to $X$ satisfy % $\M(X)$\@, and $\E{\CE{Y}{X}} = \nicefrac{1}{2}$\@.
%    \begin{align*}
%    \V{\CE{Y}{X}} &=\E{(\CE{Y}{X}-\nicefrac{1}{2})^2} 
%                  = \nicefrac{1}{2} (\nicefrac{1}{2})^2
 %                   + \nicefrac{1}{2} (-\nicefrac{1}{2})^2
%                  = \nicefrac{1}{4}
Similarly, $\CE{Y}{X_2 = 0} = \nicefrac{1}{4}$ and
$\CE{Y}{X_2 = 1} = \nicefrac{3}{4}$\@.
Hence, 
% $\E{\CE{Y}{X_2}} = \nicefrac{1}{2}$ and
$\V{\CE{Y}{X_2}} = \nicefrac{1}{16}$\@. 
Calculation
of
all 
$\V{\CE{Y}{X_u}} = \E{(\CE{Y}{X_u} - \mu)^2}$,
where $\mu \Def \E{\CE{Y}{X_u}} = \nicefrac{1}{2}$
yields the cooperation game:\\
 \begin{center}
\begin{tabular}{|c||c|c|c|c|c|c|c|c|}\hline
          & $\emptyset$ & $\{1\}$ & $\{2\}$ & $\{3\}$  & $\{1,2\}$ & $\{1,3\}$ & $\{2,3\}$ &  $\{1, 2, 3\}$ \\\hline
  $\Val{.}$ & $0$ & $0$ & $\nicefrac{1}{16}$ & $\nicefrac{1}{16}$ & $\nicefrac{1}{8}$ & $\nicefrac{1}{8}$ & $\nicefrac{1}{8}$ & $\nicefrac{1}{4}$ \\\hline
\end{tabular}
\end{center}
\vspace{2mm}
\noindent
% This game is \emph{monotonic} in that $u \subseteq v$ implies $\Val{u} \leq \Val{v}$\@.
% As a consequence, all Shapley values are nonnegative.
Using equality~(\ref{def:shapley}) the corresponding Shapley attributions are obtained.
  \begin{align*}
   \Sh{1}(\Value) &= 
     \nicefrac{1}{3} \cdot
     ( 1 \cdot (0 - 0) + 
      \nicefrac{1}{2} \cdot (\nicefrac{1}{8} - \nicefrac{1}{16}) + 
       \nicefrac{1}{2} \cdot (\nicefrac{1}{8} - \nicefrac{1}{16})   + 
       1 \cdot (\nicefrac{1}{4} - \nicefrac{1}{8})
     )
   = \nicefrac{1}{32}\\
   \Sh{2}(\Value) &=
   \nicefrac{1}{3} \cdot
   (1 \cdot (\nicefrac{1}{16} - 0) + 
    \nicefrac{1}{2} \cdot (\nicefrac{1}{8} - 0 ) + 
    \nicefrac{1}{2} \cdot (\nicefrac{1}{8} - \nicefrac{1}{16}) +
    1 \cdot (\nicefrac{1}{4} - \nicefrac{1}{8})
   )
   =   \nicefrac{3}{32}\\
    \Sh{3}(\Value) &=
    \nicefrac{1}{3} \cdot
   (1 \cdot (\nicefrac{1}{16} - 0) + 
    \nicefrac{1}{2} \cdot (\nicefrac{1}{8} - 0) + 
    \nicefrac{1}{2} \cdot (\nicefrac{1}{8} - \nicefrac{1}{16}) +
    1 \cdot (\nicefrac{1}{4} - \nicefrac{1}{8})
   )
   = \nicefrac{3}{32}
 \mbox{\@.}
   \end{align*}
Therefore, the attribution of the relative importance of the input $X_1$ for the output $Y$ is smaller then the ones for the other inputs $X_2$ and $X_3$, and the attributed relative importance of the inputs $X_2$ and $X_3$ are identical.
\end{example}
An application-specific \emph{fairness specification} needs to determine the sensitive inputs together with \emph{acceptable} upper bounds on the relative importance on the variation of the outcome. 
For example, if all the inputs $X_i$ in the example above can be considered sensitive then $X_1$ is the least biased, since the Shapley effect $\Sh{1}(\Value)$, which attributes the relative importance of $X_1$ to the decision outcome $Y$, is the smallest.
However, if this upper bound on the variance of the outcome is unacceptably high in a given social context, the decision could still be said to be \emph{unfair} with respect to the input $X_1$\@.

Fairness specifications may also be based on comparing Shapley-Owen effects. 
In particular, when specifying larger classes of fairness constraints, ratios and differences of fairness measures are relevant~\cite{feldman2015certifying,dwork2012fairness}\@.
\begin{example}
Consider a decision variable with two sensitive inputs
$X_{\mathit{gender}}$ and $X_{\mathit{ethnicity}}$\@.
For a given small $\epsilon > 0$ the constraint
    \begin{align*}
      e^{-\epsilon} < \frac{\Sh{{\mathit{gender}}}(\Value)}
                           {\Sh{{\mathit{ethnicity}}}(\Value)} < e^{\epsilon}
  \end{align*}
expresses the fact that \emph{gender} and \emph{ethnicity} have approximately the same Shapley attribution of relative importance on the decision.
\end{example}
These kinds of comparative measures of fairness are the basis of notions of fairness such as \emph{disparate impact}~\cite{feldman2015certifying} and \emph{differential fairness}~\cite{dwork2012fairness}\@.
% For some special cases,
Simplified formula for calculating \emph{differential fairness} based on ratios of Shapley values can be derived\@. 
\begin{example}\label{lem:ratio}
Let $Y = \M(X_1, X_2)$ with finite variance $\sigma^2 > 0$\@; then:
   \begin{align}
      \frac{\Sh{1}(\Value)}{\Sh{2}(\Value)} 
      &= \frac{\Val{1} + \Valtilde{2}}
              {\Valtilde{1} + \Val{2}}\mbox{\@.}
   \end{align}
\end{example}
% With $\sigma^2_{12} = \sigma^2$ and $\sigma^2_{\emptyset} = 0$ 
% we find the identity
%  $\Sh{1} 
%  = \nicefrac{1}{2} (\sigma^2_1 + \sigma^2 - \sigma^2_2) 
%  = \nicefrac{1}{2} (\sigma^2 + \Var{\CE{Y}{X_1}} - \Var{\CE{Y}{X_2}})
%  $\@.
% The statement in Lemma~\ref{lem:ratio} follows
% from a similar calculation for $\Sh{2}$\@. 

\noindent
For two inputs $\{i,j\}$ of interest the Shapley-Owen interaction index~(\ref{def:shapley.owen}) reduces to
   \begin{align}\label{eq:shapley.owen2}
   \Sh{\{i, j\}}(\Value) &= \frac{1}{(d - 1)}
   \sum_{v \,:\,v \subseteq \Compl{\{i, j\}}}
   \binom{d - 2}{|v|}^{-1} \,
 %  \frac{(d - |v| - 2)!\, |v|!}{(d - 1)!} \,
                    (\Val{v + \{i, j\}} - \Val{v + i} -  \Val{v + j}  + \Val{v})\mbox{\@.}
  \end{align}
 Equation~(\ref{eq:shapley.owen2}) coincides with the definition of two-factors interaction used in the field of \emph{Design of Experiments}~\cite{wu2015post}\@. If this term is positive than the interaction between 
 $i$ and $j$ is said to be \emph{synergistic} (i.e. profitable), 
 and if it is negative then the interaction is \emph{antagonistic}\@.
 Thus the intuition of Equation~(\ref{eq:shapley.owen2}) is that one averages this interaction index for all possible coalitions to which the subgroup $u$ belongs.
The following example illustrates Shapley-Owen effects for input sets of size $2$\@.
\begin{example}
For the game in example~(\ref{ex:prop}) we obtain from equation~(\ref{eq:shapley.owen2}) Shapley-Owen effects for subsets of inputs of cardinality two.
  \begin{align*}
  \Sh{\{1,2\}}(\Value) &=
    \nicefrac{1}{2} ((\nicefrac{1}{8} - 0 -\nicefrac{1}{16} + 0 ) + (\nicefrac{1}{4} - \nicefrac{1}{8} - \nicefrac{1}{8} + \nicefrac{1}{16})) = \nicefrac{1}{16} \\
  \Sh{\{1,3\}}(\Value) &=
    \nicefrac{1}{2} ((\nicefrac{1}{8} - 0 -\nicefrac{1}{16} + 0 ) + (\nicefrac{1}{4} - \nicefrac{1}{8} - \nicefrac{1}{8} + \nicefrac{1}{16})) = \nicefrac{1}{16} \\
 \Sh{\{2,3\}}(\Value) &=
    \nicefrac{1}{2} ((\nicefrac{1}{8} - \nicefrac{1}{16} -\nicefrac{1}{16} + 0)
    + (\nicefrac{1}{4} - \nicefrac{1}{8} - \nicefrac{1}{8} + 0)) = 0
  \end{align*}
Since the Shapley-Owen effect of the inputs $\{X_2, X_3\}$ is $0$ they have no effect on the outcome, and the underlying decision may therefore be said to be \emph{completely unbiased} with respect to this set of sensitive inputs. 
However, when considered as individual inputs, neither $X_2$ nor $X_3$ are completely unbiased, as both of their Shapley effects are non-zero, as shown in Example~\ref{ex:prop}\@.
In other words, the decision algorithm can discriminate with respect to the sensitive attributes $X_1$ and $X_2$, but it cannot discriminate with respect to both.
%Therefore, a DPLL solver might resolve to decide variable $X_1$ after, for example, resolving $X_2$ instead of $X_3$, since the relative importance of the $X_1$ and $X_2$ is bigger than the one for $X_3$ and $X_2$\@. However, such a heuristic needs to rely on an efficient approximation of the Shapley values.
\end{example}

\subsection{Fair Attribution}
The Shapley value is an instance of a larger class of 
\emph{probabilistic values}~\cite{roth1988shapley}\@,
but it is the only one satisfying the following four conditions~\cite{shapley1953value} for all
possible values $\Value$\@:
\begin{enumerate}
\item \label{Sh.efficiency} (Efficiency) $\sum_{i=1}^d \Sh{i}(\Value) = \Val{[1,d]}$\@.
\item \label{Sh.symmetry} (Symmetry) If $\Val{u + i} = \Val{u + j}$ for all $u \subseteq \Compl{\{i, j\}}$ then $\Sh{i}(\Value) = \Sh{j}(\Value)$\@.
\item \label{Sh.null} (Dummy) If $\Val{u + i} = \Val{u}$ for all $u \subseteq [1,d]$ then $\Sh{i}(\Value) = 0$\@.
\item \label{Sh.additivity}  (Additivity) $\Sh{i}(\Value_1 + \Value_2) = \Sh{i}(\Value_1) + \Sh{i}(\Value_2)$ for all $i \in [1,d]$\@,
\end{enumerate}
where the value $(\Value_1 + \Value_2)(u)$ adds 
the two values $\Value_1(u)$ and $\Value_2(u)$\@.
Additivity also implies that multiplying $\Value$ by a scalar multiplies all of the Shapley values by that same scalar; that is: $\Sh{i}(c \cdot \Value) = c \cdot \Sh{i}(\Value)$\@.
% In addition, if the game $\Value$ is \emph{monotonic}, 
% that is, $u \subseteq v$ implies $\Val{u} \leq \Val{v}$, then
% $\Sh{i}(\Value) \geq 0$ for all $i \in D$\@.
Informally, \emph{efficiency} is interpreted as stating that the variance of the outcome can be explained in terms of attributed relative importance of each input. 
Moreover, the \emph{dummy} property states that attributions to inputs that do not add to the relative importance of other inputs should have attributed zero importance, 
whereas \emph{symmetry} demands that
two inputs who contribute equally to the relative importance of other inputs should be indistinguishable in that they have identical attribution. 
Alternatives to Shapley's original axiomatization  
have been developed~\cite{myerson1980conference,hart1989potential,young1985monotonic}\@.
For example, the axioms (Dummy), (Symmetry), and (Additivity) can be replaced by the equivalent single 
\emph{(Balanced Contributions)} axiom~\cite{myerson1980conference,hart1989potential}\@:
\begin{align}
\Sh{i}(\Value) - \Sh{i}(\Restrict{\Value}{\Compl{j}})
&=
\Sh{j}(\Value) - \Sh{j}(\Restrict{\Value}{\Compl{i}})\mbox{\@,}
\end{align}
where $\Restrict{\Value}{\Compl{i}}$ is the restriction of the 
game $\Value$ to the inputs $X_{\Compl{i}}$\@.
Balanced contribution is a principle 
of fairness in cooperation, as it states that every pair of players should share equally in the gain of their cooperation.

More generally, Shapley-Owen values are based on game-theoretic axioms that differ slightly from the ones of the
Shapley value~\cite{grabisch1999axiomatic}\@.
\begin{enumerate}
    \item (Additivity) $\Sh{u}(\Value_1 + \Value_2) = \Sh{u}(\Value_1) + \Sh{u}(\Value_2)$ for all $u\subseteq [1,d]$\@.
    \item (Dummy) If $\Val{u + i} = \Val{u}$ for all $u \subseteq \Compl{i}$ then $\Sh{i}(\Value) = \Val{i}$ and
    $\Sh{v + i}(\Value) = 0 $ for every $\emptyset \neq v \subseteq \Compl{i}$\@.
    \item (Symmetry)
      For all games $\Value$ and permutations 
      $\sigma \in \Pi([1,d])$\@:
      $\Sh{u}(\Value) = \Sh{\sigma(u)}(\Value_\sigma)$\@,
      where the permutation game $\Value_\sigma$ maps the direct image $\sigma(u) \Def \{ \sigma(i) \,|\, i \in u\}$ to 
      $\Val{u}$\@. 
    \item (Recursivity) 
      For every $\Value$, 
      any $u \subseteq [1,d]$ such that $|u| > 1$, 
      any $i \in u$\@,
      and the games $M_i(v) \Def \Val{v \cup \{i\}} - \Val{v}$\@:
       %\begin{align*}
       $\Sh{u}(\Value) =
            \Sh{u\setminus\{i\}}(M_i)
          - \Sh{u\setminus\{i\}}(\Restrict{\Value}{\Compl{i}})$\@. 
       %\end{align*}
    %  for the game $M_i(v) \Def \Val{v \cup \{i\}} - \Val{v}$\@.
      % and the restricted $\Restrict{\Value}{\Compl{i}}$ has been defined above\@.
\end{enumerate}
The Shapley-Owen interaction index~(\ref{def:shapley.owen}) satisfies these axioms, and it is unique in that it is the only such attribution which restricts to singleton Shapley values~\cite{grabisch1999axiomatic}\@.
Recursivity says that the interaction
of the players in $u$ is equal to the interaction between the players in $u\setminus \{j\}$ in
the omnipresence of $j$, 
minus the interaction between the players of 
$u\setminus \{j\}$ in the absence of $j$\@.
Overall, if we accept these "fairness" conditions on attributions, then variance-based Shapley-Owen effects are the only possible "fair" attribution of the relative importance of inputs and their interactions.

\section{Polynomial Chaos Expansion}\label{sec:pce}

Polynomial chaos expansion (PCE) is a \emph{spectral decomposition} that expands the model output to an infinite series of orthogonal polynomials in random model inputs.
Assume as given a complete orthonormal polynomial basis
$(\Psi_\alpha(x))_{\alpha \in \Nat}$ of $L_2(\Omega)$\@;
% where $\Psi_\alpha \in \Real[x_1, \ldots, x_d]$\@.
therefore:\footnote{
$\delta_{\alpha\beta}$ is the Kronecker symbol.} 
   \begin{align}\label{eq:orthonormal.expectation}
       \E{\Psi_\alpha(X)} &= 0 &\text{for~}\alpha \neq 0 \\
       % \E{\Psi_{\alpha}(X) \Psi_{\alpha}(X)} &=
   %    \V{\Psi_{\alpha}(X)} &= 1    \\
        \label{eq:orthonormal.variance}
       \E{\Psi_\alpha(X)\,\Psi_\beta(X) } &= \delta_{\alpha,\beta}\mbox{\@.}  %  & \text{for~}\alpha \neq \beta 
   \end{align}
% for the Kronecker symbol $\delta_{\alpha,\beta}$\@.
The PCE for the square-integrable random variable $\M(X)$ is given by
   \begin{align}\label{def:pce}
       Y \Def \M(X) = \sum_{\alpha=0}^\infty y_\alpha \Psi_\alpha(X)\mbox{\@.}
   \end{align}
$y_\alpha \in \Real$ is said to be the $\alpha$-th PCE coefficient.
We also use multi-indices $\alpha \in \Nat^d$ as summation indices in the PCE, in a slight abuse of notation.
We therefore assume a fixed enumeration of the
multi-indices $\alpha \in \Nat^d$ always starting with the multi-index $0$\@. 
The coefficients $y_\alpha$ of the PCE in~(\ref{def:pce})
are determined by the identity
\begin{align}\label{Galerkin}
    \E{Y\, \Psi_{\alpha}(X)} = \sum_{\beta = 0}^\infty y_\beta\, \E{\Psi_{\beta}(X) \Psi_{\alpha}(X)}
                    = \sum_{\beta = 0}^\infty y_\beta \delta_{\alpha\beta}
                    = y_\alpha\mbox{\@,}
\end{align}
since $\E{\Psi_{\alpha}(X) \Psi_{\beta}(X)} = \delta_{\alpha\beta}$ by orthonormality\@.
The expectation $\E{Y\, \Psi_{\alpha}(X)}$ is also said to be a {\em Galerkin projection}\@.
Now, by linearity of expectation and identity~(\ref{eq:orthonormal.expectation}), 
the expectation of $\M(X)$ coincides with the initial Galerkin 
projection in the PCE of $\M(X)$\@.
\begin{align}\label{pce:expectation}
    \mu \Def \E{\M(X)} &= y_{0}
\end{align}
Bessel's inequality implies that the infinite sum
$\sum_{\alpha = 0}^\infty y_{\alpha}^2$
of squared Galerkin projections converges
and it satisfies the inequality
$\sum_{\alpha = 1}^\infty y_{\alpha}^2
 \leq \sigma$\@.
 As a consequence, the Parseval identity, together with the identity~(\ref{eq:orthonormal.variance}), gives the convergence of the PCE~(\ref{def:pce}) and the characterization of the variance of $\M(X)$
 in terms of the coefficients of the PCE for $\M(X)$\@.
\begin{align}\label{pce:variance}
  \sigma^2 &\Def \V{\M(X)} = \sum_{\alpha = 1}^\infty y_{\alpha}^2\mbox{\@.}
\end{align}
Moreover, convergence of the PCE~(\ref{def:pce}) is known to be exponential in many cases~\cite{wiener1938homogeneous,cameron1947orthogonal,xiu2002wiener}\@.

%Without loss of generality, we may assume that the 
% polynomial $\Psi_\alpha(X)$ has degree $\alpha$\@.

Determining the PCE~(\ref{def:pce}) for a given $\M(X)$ proceeds by (1) constructing an orthonormal polynomial basis with respect to the input distribution of $M$, and (2) determining the corresponding PCE coefficients.

\subsection{Orthonormal Basis}

The family of Legendre (Hermite, Jacobi, Laguerre) polynomials are orthogonal with respect to the uniform (Gaussian, $\beta$, $\Gamma$) distribution over the domain $[-1,1]$ ($\Real$, $[-1,1]$, $[0,\infty)$)~\cite{xiu2002wiener}\@. 
This correspondence extends to discrete distribution in 
that Charlier (Krawtchouk, Meixner, Hahn) polynomials
are orthogonal with respect to the Poisson (binomial, negative binomial, hypergeometric) distribution over the discrete domain $\Nat$ ($[0,n] \cap \Nat$, $\Nat$, $[0,n] \cap \Nat$)\@.
But there are also many commonly used
probability distributions, such as L\'evy, Weibull, and $\chi$-square, for which exact recurrences are not known~\cite{gautschi2004orthogonal}\@.

If input independence is assumed then
multivariate orthogonal polynomial basis vectors
$\Psi_\alpha(x)$, 
for $ \alpha = \alpha_1\ldots\alpha_d$,
are assembled by the product 
\begin{align}\label{multivariate.independent}
    \Psi_\alpha(x) &\Def \prod_{i=1}^d P_{\alpha_i}(x_i)\mbox{\@,}
\end{align}
of the univariate $(P_{\alpha_i}(x_i))_{\alpha_i \in \Nat}$ of degree $\alpha_i$, for $i = 1, \ldots d$,
which are orthogonal in the sense that
$\E{P_{\alpha_i}(X_i)P_{\alpha_j}(X_j)} = 0$
for $i \neq j$\@.
Orthogonality $\E{\Psi_\alpha(X)\Psi_\beta(X)} = \delta_{\alpha\beta}$ of the multivariate polynomials  $\Psi_\alpha(X)$ in~(\ref{multivariate.independent}) follows from input independence and orthogonality of their univariate counterparts.
%\begin{example}
For the univariate \emph{Legendre polynomials} $\Le{\alpha_i}(x_i) \in \Real[x_i]$, $i = 1, \ldots, d$, of degree $\alpha_i \in \Nat$, for example, 
the Legendre polynomial $\Le{\alpha}(x) \in \Real[x]$
of multi-degree $\alpha= (\alpha_1, \ldots, \alpha_d)$ are defined to be
\begin{align}\label{legendre}
    \Le{\alpha}(x) &\Def \prod_{i=1}^d \Le{\alpha_i}(x_i)\mbox{\@.}
\end{align}
Let $\widetilde{\Le{\alpha}}(x)  \Def \frac{\Le{\alpha}(x)}{\Norm{\Le{\alpha}(x)}{f}}$ 
be the normalized Legendre polynomial of multi-degree $\alpha$\@.
% Normalized multivariate \emph{Hermite polynomials} $\widetilde{\He{\alpha}}(x)$ are defined similarly, and a PCE of the form $\sum_{\alpha \in \Nat^d} y_\alpha \widetilde{\He{\alpha}}(X)$ is known as a 
% \emph{Wiener-Hermite expansion}~\cite{wiener1938homogeneous}\@.
%\end{example}
The case of dependent input variables is sligthly more complicated~(Appendix~\ref{app:dependent.PCE})\@.

\subsection{PCE Coefficients}

By identity~(\ref{Galerkin}) the computation of the PCE coefficients is reduced to the calculation of an expectation. In some special cases PCE coefficients are obtained analytically~(for example, ~\cite{sullivan2015introduction}, Lemma 11.8)\@.
% \begin{example}[(\cite{sullivan2015introduction}, Lemma 11.8)]
% For example, let $\Xi \sim \mathcal{N}(0,1)$ be a standard Gaussian random variable, 
%and $\He{k}$ the (univariate) Hermite polynomial of degree $k \in \Nat$\@. 
%If $\M(x) \Def \sum_{k=0}^\infty a_k x^k$ is any convergent power series then $\E{\M(\Xi) \,\He{k}(\Xi)} = k!\,a_k$ 
%for all $k \in \Nat$~(\cite{sullivan2015introduction}, Lemma 11.8)\@.
% \end{example}
In general, however, standard numerical methods such as \emph{Gaussian quadrature} are used to approximate the Galerkin projection~(\ref{Galerkin})\@:
\begin{align}\label{Gaussian.quadrature}
  y_\alpha 
  = \E{\M(X) \Psi_{\alpha}(X)}
  = \int_{\Omega} \M(x) \Psi_\alpha(x) f(x)\,dx
  \approx \sum_{k = 0}^N \Sample{w}{k} \M(\Sample{x}{k}) \Psi_\alpha(\Sample{x}{k})\mbox{\@.}
  \end{align}
The  quadrature points $\Sample{x}{k}$ and the set of weights $\Sample{w}{k}$\@, for $k = 0, \ldots, N$, are derived from polynomial interpolation. 
The Golub-Welsch algorithm requires $\mathcal{O}(N^2)$ operations~\cite{golub1969calculation}\@.
% , and asymptotic formulas for large $N$ require $\mathcal{O}(N)$ operations. 
% and guarantees exactness in the evaluation of the integrals of polynomials. 
% ~\cite{gander2000adaptive}; 
% see also~\cite{golub1969calculation,ryu2015extensions}\@.
% Identities for the error of a Gaussian quadrature rule % involve the $(2N+2)$-th derivation of the weight $f$\@.
% Stoer, Belirsch, 2002, Thm 3.6.24
% Kahaner, Moler, NAsh, 1989, Chapter 5.2
Error estimates for Gaussian quadrature have been studied
extensively~\cite{stoer1980introduction,kahaner1989numerical}\@.
The $N$-point Gauss quadrature formula~(\ref{Gaussian.quadrature}) has order of accuracy exactly $2N+1$,
that is, it is exact for every polynomial of degree $2N+1$ or lower,
and no quadrature formula has order of accuracy higher than this~\cite{gander2000adaptive,ryu2015extensions}\@.
Standard multivariate Gaussian quadrature is achieved by a tensor product of univariate integration rules.
Therefore, the selection of a maximal polynomial degree $p$ requires $(p + 1)$ integration points in each dimension, leading to
$N = (p + 1)^d$ in equation~(\ref{Gaussian.quadrature})\@.
\emph{Sparse quadrature rules}~\cite{trefethen2008gauss} such as \emph{Smolyak's quadrature}~\cite{smolyak1963quadrature}
use considerably fewer nodes at the cost of some accuracy\@. 
%
% The degree of exactness is $2N-1$ for $N$ the number of quadrature points. 
% It is the largest possible degree of exactness.
% An \emph{a posteriori} error estimate of the Gaussian quadrature error in the estimation of the PCE
% coefficients in equation~(\ref{Gaussian.quadrature}) can be calculated by taking the expectation value of the residual
%mean-square error % $\E{\M(X) − \M^{\mathit{PC}}(X)}$ 
%by integrating it with the same quadrature rule
%and on the same nodes~(\cite{UQdoc_21_104}, page 10)\@.
Other efficient methods for determining PCE coefficients include \emph{least-squares regression}~\cite{berveiller2006stochastic}, 
\emph{least-angle regression}~\cite{blatman2011adaptive}
to favour sparsity in high dimension, and
\emph{orthogonal matching pursuit}~\cite{pati1993orthogonal}\@.
Although the PCE coefficients are usually only obtained numerically, for the sake of readability and to avoid distracting form the main path of the developments we make the simplifying assumption that the corresponding approximation errors are negligible. 
However, adding these effects to the error analysis below is a standard exercise.

% \subsection{Moments}
% The moments $\mu^k$ of $Y = \M(X)$, for $k \in \Nat$ can be analytically expressed in terms of its PCE.
% \begin{align}
%    \mu^k &\Def \int_{\Omega} \M^k(x) f(x) \, dx 
%           = \int_{\Omega} (\sum_{i=1}^\infty y_i \Psi_i(x))^k f(x) \, dx\mbox{\@.}
%\end{align}
%By applying the orthogonal identities of multivariate polynomials of PCE, the analytical expressions of the first- and second-order moments are attained as follows.
%  \begin{align}
%      \mu =\E{Y} &= y_0 \\
%      \sigma^2= \Var{Y} &= \sum_{\alpha \in \Nat^d \setminus \{0\}}^\infty y_{\alpha}^2\mbox{\@.} 
%  \end{align}
% Unlike the first two order moments, the higher order moments do not have compact expressions.
%% Since a (truncated) PCE is a polynomial in which the random variables of all terms are separable, 
%% its powers are also polynomials with separable random variables. 
%% Because of this fact, the higher order moments of the model's responses is expressable in terms of the moments of the univariate distributions associated with the input random variables.
%An approximation of the pdf of the outcome $Y = \M(X)$ is obtained by using the PCE %as a \emph{stochastic response surface} and by plotting the obtained sample set %using histograms or kernel density estimators.

\subsection{Truncated PCE}\label{subsec:truncated.PCE}

For practical purposes, the PCE is truncated
as $\M^l(X) \Def \sum_{\alpha=0}^l y_\alpha \Psi_{\alpha}(X)$ for some $l \in \Nat$\@.
We always assume, without loss of generality, that the initial term of a truncated PCE is the expectation $\mu$ of $\M(X)$\@.
Now, the truncation error
   \begin{align}\label{def:truncation.error}
    \epsilon_l &\Def \M(X) - \M^l(X)
   \end{align}
is orthogonal to the truncated PCE $\M^l(X)$~\cite{cameron1947orthogonal,xiu2010numerical} (Appendix~\ref{app:truncation.error})\@.
Therefore, by the Pythagorean theorem for inner product spaces,
$\Norm{\M(X)}{f}^2
=\Norm{\epsilon_l + \M^l(X)}{f}^2  
= \Norm{\epsilon_l}{f}^2 + \Norm{\M^l(X)}{f}^2
% = \Norm{\M(X) - \M^l(X)}{f}^2 + \Norm{\M^l(X)}{f}^2
$\@.
% Assuming input independence, 
Variance decomposition~(\ref{pce:variance}) applies, and 
the truncation error $\epsilon_l$ is characterized by the identity
  \begin{align}\label{error.identity}
     \V{\epsilon_l} &= \sigma^2 - \sum_{\alpha = 1}^l y^2_{\alpha}\mbox{\@.}
  \end{align}
The truncation errors $\V{\epsilon_l}$ are nondecreasing as $l$ goes to infinity, and strictly decreasing under the reasonable assumption that $y_{\alpha} \neq 0$,
and $\lim_{l\to\infty}(\V{\epsilon_l}) = 0$\@.
The identity~(\ref{error.identity}) serves as a stoppage criterion in the iterated computation of the truncated PCE $\widehat{\M}_l$\@.
\begin{proposition}\label{prop:stoppage.criterion}
For $\M(X)$ with variance $\sigma^2 < \infty$,
the truncation
$\widehat{\M}_l(X) \Def \sum_{\alpha=0}^l y_{\alpha} \Psi_{\alpha}(X)$ 
of the PCE of $\M(X)$\@,
and $\omega > 0$\@:
$\Var{\epsilon_l}< \omega$ if and only if 
$\sigma^2 - \sum_{\alpha=1}^l y_{\alpha}^2 < \omega$\@. 
\end{proposition}
The expection of the truncation error is zero, since
 $\E{\epsilon_l} = \E{\M(X)} - \E{\widehat{\M}_l(X)} = y_0 - y_0 = 0$\@.
From the identity (\ref{error.identity}) 
and Chebyshev's inequality one therefore 
obtains confidence intervals for the preciseness of the truncated PCE\@.
\begin{proposition}\label{prop:chebishev}
For $\M(X)$ with variance $\sigma^2 < \infty$,
the truncation
$\sum_{\alpha=0}^l y^2_{\alpha} \Psi_{\alpha}(X)$ of the PCE of $\M(X)$\@,
and real-valued $t > 0$\@:
\begin{align}
    \E{|\epsilon_l| \geq t} &\leq \nicefrac{1}{t^2}
    (\sigma^2 - (\sum_{\alpha=1}^l y_{\alpha}^2))^2
    \mbox{\@.}
\end{align}
% where $\Var{\epsilon_l}$ is calculated as identity~
% from $\sigma^2$ and the truncated PCE $\hat{\M}_l(X)$\@.
\end{proposition}
In engineering problems only low-order interactions between the input variables are typically relevant.
This observation is known as the \emph{sparsity-of-effects}~\cite{luethen2021sparse}\@. 
In these cases, 
% the truncation scheme $\sum_{i = 1}^d \chi_{\alpha_i > 0} \leq p$ restricts the maximum interactions to a given $p \in \Nat$\@. 
% The 
the \emph{$q$-norm}~\cite{blatman2009adaptive} truncation scheme % $\Norm{\alpha}{q}  \Def 
$(\sum_{i = 1}^d \alpha_i^q)^{\nicefrac{1}{q}}\leq p$
for a multi-index $\alpha = (\alpha_1, \ldots, \alpha_d)$
corresponds, for $q = 1$, exactly to the total degree of $\alpha$\@. 
For  $q < 1$, it includes all the univariate high-degree terms, but excludes high-degree terms with many interacting variables.
The construction of sparse and sufficiently 
accurate PCE approximation is an art in itself~\cite{babacan2009bayesian,dai2009subspace,blatman2009adaptive,blatman2010adaptive,blatman2011adaptive,shao2017bayesian}\@.
For example, sparse PCEs are constructed in the Bayesian framework using the Kashyab information criterion for model selection~\cite{shao2017bayesian}\@. 
% \emph{subspace pursuit} sparse regression algorithm~\cite{dai2009subspace}, or
% \emph{Bayesian compressive sensing}~\cite{babacan2009bayesian}\@.

\subsection{Sparse PCE}\label{subsec:sparse.PCE}

The identity~(\ref{error.identity}) on the 
truncation error together with the \emph{sparsity-of-effects} hypothesis
suggests a simple approximation algorithm with input parameters $q \in \Real$ and approximation bound $\epsilon > 0$\@.
The algorithm starts with the initial approximation
$\widehat{M}_0(X) \Def y_0 (= \E{\M(X)})$,
and iteratively adds new PCE terms $y_{\alpha} \Psi_{\alpha}(X)$ such that $\alpha$ satisfies the $q$-norm in the $p$-th but not in the $(p-1)$-th iteration step.
The algorithm stops when the norm of the error terms falls below $\epsilon$ according to Proposition~\ref{prop:stoppage.criterion}\@.
This iteration terminates with $\Norm{\M(X) - \widehat{\M}_l(X)}{f} < \epsilon$, where $l$ is the number of steps in the iteration\@.
Alternatively, the probabilistic stoppage criterion in Proposition~\ref{prop:chebishev} can be used, in which case we get a confidence interval for the error.
Now, the sparsity of the PCE approximation $\widehat{\M}_l(X)$ is obtained by 
adding the terms $y_{\alpha} \Psi_{\alpha}(X)$ to the current PCE approximation 
only if $y_{\alpha}$ is sufficiently large.
Otherwise, $y_{\alpha}$ is added to a \emph{waiting} set,
and the PCE terms are chosen in every iteration as the maximum absolute value of the coefficients in the \emph{waiting} set and the current $y_{\alpha}$\@.
If the maximum error $|y_{\alpha} - \widehat{y}_{\alpha}|$ for the computed approximations $\widehat{y}_{\alpha}$ of the PCE coefficients $y_{\alpha}$ is bounded by some $\kappa > 0$, then
that bound is used, for example, to decide whether $\widehat{y}_{\alpha}$ is put into the
\emph{waiting} set or not.
This additional check prevents terms with a coefficient of $\widehat{y}_{\alpha}$, whose real coefficient $y_{\alpha}$ is zero or close to zero, from being prematurely added to the truncated PCE.

\subsection{Sensitivity}

Sensitivity indices for $\M(X)$, such as the Sobol' indeces~(\ref{def:sobol}) and~(\ref{def:sobolT}), are obtained analytically 
from the coefficients of the PCE of $\M(X)$\@.
In a first step, the summands of the PCE~(\ref{def:pce}) are rearranged as
\begin{align}\label{pce.rearrange}
    \M(X) & = \sum_{u \subseteq [1,d]} \sum_{\alpha=0}^\infty y_\alpha \,\pi_u(\Psi_\alpha(X))\mbox{\@,}
\end{align}
where the characteristic function $\pi_u(\Psi_\alpha(X))$ holds if and only if $X_u$ are the variables occurring in $\Psi_\alpha(X)$\@.
% $\mathcal{A}_u$ contains exactly the indices $\alpha \in \Nat^d$ for which the subset $X_u$ of inputs is exactly the set of indeterminates occurring in $\Psi_{\alpha}$\@.
% $\mathcal{A}_u$ contains exactly the degrees $\alpha \in \Nat^d$ for which $\alpha_k \neq 0$ if and only $k \in u$\@. 
Because the inner summands in~(\ref{pce.rearrange}) depend only on the variables $X_u$\@, 
the partial effects $\M_u(X_u)$ of $\M(X)$
and their variances $\sigma^2_u$
are defined as
\begin{align}\label{sum.of.effects}
       \M_u(X_u) &\Def \sum_{\alpha=0}^\infty y_\alpha \,\pi_u(\Psi_\alpha(X)) \\
       \label{def:variance.partial}
        \sigma^2_u &\Def \V{\M_u(X_u)}\mbox{\@.}
   \end{align}
The expectation $\E{\M_u(X_u)}$ of the partial effects $\M_u(X_u)$ equals $0$ as a result of 
identity~(\ref{eq:orthonormal.expectation})\@.
Moreover,  by identity~(\ref{eq:orthonormal.variance}), 
the random variables $\Psi_\alpha(X)$ are independent.
Therefore the variance $\sigma_u^2$ of the partial effects
$\M_u(X_u)$ is expressed in terms of a subset of the PCE 
coefficients for $\M(X)$\@.
   %\begin{align}\label{variance.total.effects}
   %    \sigma^2_u \Def \V{\M_u(X_u)} & 
   %    = 
   %     \sum_{\alpha \in \mathcal{A}_u \setminus \{0\}} y_\alpha^2 \,\V{\Psi_{\alpha}(X)}
   %   = \sum_{\alpha \in \mathcal{A}_u \setminus \{0\}} y_\alpha^2\mbox{\@.}
   %\end{align}
    \begin{align}\label{variance.total.effects}
       \sigma^2_u = \V{\M_u(X_u)} & 
       = 
        \sum_{\alpha = 1}^\infty y_\alpha^2 \,\V{\pi_u(\Psi_{\alpha}(X)})
      = \sum_{\alpha =1}^\infty y_\alpha^2 \,\chi_u(\Psi_{\alpha}(X))\mbox{\@.}
   \end{align}
The right-most identity holds, since the $\Psi_{\alpha}(X)$ are assumed to be normalized~(\ref{eq:orthonormal.variance})\@.
Using the shorthand $\chi_u(\alpha)$ for $\chi_u(\Psi_{\alpha}(X))$\@, the variance of the partial effects of $\M(X)$ is obtained analytically from the PCE of 
$\M(X)$ as follows.
\begin{lemma}\label{lem:partial.variance.PCE}
For the PCE $\M(X) = \sum_{i=0}^\infty y_\alpha \Psi_{\alpha}(X)$, 
the partial effects $\M_u(X_u)$~(\ref{sum.of.effects})  of $\M(X)$\@,
and their respective variances $\sigma^2_u$~(\ref{def:variance.partial})\@:
\begin{align}\label{eq:partial.variance.PCE}
    \sigma^2_u &= 
    \sum_{\alpha =1}^\infty y_\alpha^2 \,\chi_u(\alpha)\mbox{\@.}
\end{align}
\end{lemma}
We are now analyzing the approximation error of computing the variances $\sigma_u^2$ from a truncated PCE\@.
Hence, let $\widehat{M}_l(X) \Def \sum_{i = 0}^l y_{\alpha} \Psi_{\alpha}(X)$ be a truncated PCE for $\M(X)$\@,
and 
\begin{align}\label{approximated.variance.total.effects}
       \widehat{\sigma}^2_u  &\Def \sum_{\alpha =1}^l y_\alpha^2 \chi_u(\alpha)\mbox{\@.}
   \end{align}
be an approximation of the variances $\sigma^2_u$ in~(\ref{variance.total.effects})\@.
Now, 
  \begin{align}
  0 \leq \sigma^2_u - \widehat{\sigma}^2_u  = \sum_{\alpha = l+1}^\infty y^2_{\alpha} \chi_u(\alpha) \leq \sum_{\alpha = l+1}^\infty y^2_{\alpha} = \epsilon_l\mbox{\@,}
  \end{align}
where $\epsilon_l$ is the truncation error~(\ref{def:truncation.error})\@.
This inequality and the identity~(\ref{error.identity}) for the truncation error $\epsilon_l$ gives a uniform upper bound for the approximation error of the variances of the partial effects.
\begin{corollary}\label{prop:approx.partial.variances}
For $\M(X)$ with variance $\sigma^2 < \infty$\@, 
$\widehat{M}_l(X) \Def \sum_{i = 0}^l y_{\alpha} \Psi_{\alpha}(X)$ a truncation of the PCE for $\M(X)$\@,
and $\widehat{\sigma}^2_u(l)$ an approximation of $\sigma^2_u$ as defined in~(\ref{approximated.variance.total.effects})\@: 
   \begin{align}
       |\sigma^2_u - \widehat{\sigma}^2_u(l)| &\leq \sigma^2 - \sum_{i=1}^l y_{\alpha}^2\mbox{\@.}
   \end{align}
In addition, $\lim_{l \to \infty} |\sigma^2_u - \widehat{\sigma}^2_u(l)| = 0$\@.
\end{corollary}
In case the inputs in $X$ are independent, 
the Hoeffding decomposition~(\ref{hoeffding})
of $\M(X)$ exists and is unique (Appendix~\ref{app:anova})\@. 
In this case, the $\M_u(X_u)$ in~(\ref{sum.of.effects}) are exactly the summands~(\ref{hoeffding}) of the Hoeffding decomposition; therefore, $\E{\M_u(X_u)\M_v(X_v)} = \delta_{uv}$\@,  
and the  Sobol' indeces (\ref{def:sobol}) and total Sobol' indeces~(\ref{def:sobolT}) 
of any order are obtained,
from identity~(\ref{eq:partial.variance.PCE}), analytically from the coefficients of the PCE for $\M(X)$\@. 
This fact has originally been observed in~\cite{sudret2008global}\@.
Similarly, the analysis of variance~(\ref{ANOVA}) is obtained 
from the identities~(\ref{pce.rearrange}) and~(\ref{eq:partial.variance.PCE}) in case of independent inputs.

\subsection{PCE for Dependent Inputs}

\newcommand{\T}{\mathcal{T}}

The Rosenblatt transformation method involves mapping the original random variables $X = (X_1, \ldots, X_d)$ to the independent and uniformly distributed $U = (U_1, \ldots U_d)$ on the hypercube $[-1,1]^d$\@.
A Rosenblatt~\cite{rosenblatt1952remarks} transform $\T: \Real^d \to \Real^d$ is defined by
\begin{align}
    \T(x_1, \ldots, x_d) &\Def (F_k(x_k \,|\,x_{[1,k-1]}))_{k\in [1,d]}\mbox{\@,}
\end{align}
with $F_k$, for $k \in [1,d]$, the marginal cdf 
   $F_k(x_k \,|\,x_{[1,k-1]}) \Def \Prob{X_k \leq x_k \,|\, X_1 = x_1, \ldots, X_{k-1} = x_{k-1}}$\@.
These marginals are obtained analytically from the pdf $f$ of $X$ (Appendix~\ref{app:marginal.cdfs})\@.
Note that the Rosenblatt transform is not unique, and that there are $d!$ different ways to define such a transform, depending on which variable is chosen in each iteration. 
Now, for $U = \T(X)$, the $U_i$ are independent and uniformly distributed on $[-1,1]$\@.
In addition, if the inverses of the marginal cdfs exist,
then
\begin{align}
\T^{-1}(u_1, \ldots, u_d) &\Def (F_k^{-1}(u_k))_{k \in [1,d]}
\end{align}
is the inverse of the Rosenblatt transform.
Therefore, $X = \T^{-1}(U)$ and the pdf of $X$ is $f$\@.
\begin{lemma}\label{lem:general.PCE}
Let $\M(X)$ be square-integrable,
$U=(U_1, \ldots, U_d)$ be a $d$-dimensional vector of independent and uniformly distributed random variables, and $\T^{-1}$ with $X = \T^{-1}(U)$\@.
Then the PCE of $\M(X)$ is obtained as
\begin{align}
    \M(X) = (\M \circ \T^{-1})(U)
          = \sum_{\alpha \in \Nat^d} y_{\alpha} \widetilde{\Le{\alpha}}(U)\mbox{\@,}
\end{align}
where $\widetilde{\Le{\alpha}}(U)$~(\ref{legendre}) is the normalized Legendre polynomial of multi-degree $\alpha$, and the coefficients $y_\alpha$ are determined by the identity
  \begin{align}
  y_{\alpha} &= \E{(\M \circ \T^{-1})(U)\,\widetilde{\Le{\alpha}}(U)}\mbox{\@.}
  \end{align}
\end{lemma}
We would get the Wiener-Hermite~\cite{wiener1938homogeneous} expansion if we had chosen an isoprobabilistic transformation into Gaussian instead of uniformly distributed independent variables.
% Generally, under some mild restrictions, there is an isoprobabilistic transformation into any other distribution.

\section{Computing Shapley-Owen Effects}\label{sec:computing.soe}

We describe the spectral decomposition of Shapley-Owen effects in terms of the PCE of the underlying decision
model $\M(X)$\@.
Let $\T$ be a Rosenblatt transform with $U = \T(X)$ and inverse $\mathcal{T}^{-1}$ such that $U = \mathcal{T}^{-1}(X)$\@.
We also assume as given a PCE of $(\M \circ \T^{-1})(U)$ as described in Lemma~\ref{lem:general.PCE}\@.
The value functions of relative importance for the input vector $X$ and the independently and uniformly distributed variables $U = \T(X)$ are defined, for $u \subseteq [1,d]$\, as
\begin{align}
   \Value^X(u) &\Def \V{\CE{\M(X)}{X_u}} \\
    \Value^U(u) &\Def \V{\CE{(\M \circ \T^{-1})(U)}{U_u}}\mbox{\@.}
\end{align}
These values are consistent with the Sobol'
indices, but only $\Value^U$ can be decomposed with the identity~(\ref{def:sobol}), because the random variables in $U$ are independent by construction\@.
From these definitions, Shapley-Owen effects are invariant under the inverse Rosenblatt transformation.
\begin{lemma}\label{lem:invariant}
For all $u \in [1,d]$\@:
\begin{align}
 \Sh{u}(\Value^X)  &= \Sh{u}(\Value^U)\mbox{\@.}
\end{align}
\end{lemma}
The identity in Lemma~\ref{lem:invariant} holds, because of the defining equality~(\ref{def:shapley.owen}) of Shapley-Owen values and the identity
    $\V{\CE{\M(X)}{X_v}} 
    = \V{\CE{\M(\T^{-1}(U))}{U_v}}$\@.
The statement in Lemma~\ref{lem:invariant} also holds for other iso-probabilistic transforms~\cite{lebrun2009rosenblatt}\@.
The decision of which transform to use is problem-specific and depends strongly on the available information about the input distribution~\cite{mara2021polynomial}\@.

Using the identities~(\ref{def:sobol})
and~(\ref{variance.total.effects}), 
the value $\Value^U(v)$ is expressed in terms of the coefficients of the PCE for 
$\M(X) = (\M \circ \T^{-1})(U)$\@:
  \begin{align}\label{eq:value.PCE.coefficients}
    \Value^U(u) 
  = \sum_{v:\,v \subseteq u} \sigma^2_v 
  = \sum_{v:\,v \subseteq u} \sum_{\alpha =1}^\infty y^2_{\alpha} 
  \chi_v(\alpha)
  = \sum_{\alpha =1}^\infty y^2_{\alpha} \sum_{v:\,v \subseteq u}\chi_v(\alpha)\mbox{\@,}
\end{align}
where $\sigma_v^2 \Def \V{\M \circ \T^{-1})_v(U_v)}$\@.
Identity~(\ref{eq:value.PCE.coefficients}) motivates the definition, for $\alpha \neq 0$,
of the value function
\begin{align}\label{def:value.alpha}
\Value^\alpha(u) &\Def \sum_{v:\,v \subseteq u}\chi_v(\alpha)\mbox{\@,}
\end{align}
such that $\Value^U(u)$, and by Lemma~\ref{lem:invariant} also $\Value^X(u)$, is a linear combination of $\Value^\alpha(u)$\@.
$\Value^\alpha$ is a cooperation game, since, by definition, 
$\Value^{\alpha}(\emptyset) = 0$\@. 
It is also easy to see that $\Value^{\alpha}(u) \in \{0, 1\}$\@. 
Therefore, for any $v$, the absolute value of the inner sum in the definitional identity~(\ref{def:shapley.owen}) of Shapley-Owen values for $\Value^\alpha$ is bounded from above:
\begin{align}\label{bound}
|\sum_{w \,:\,w \subseteq u} (-1)^{|u| - |w|} \Value^{\alpha}(v + w)|
&\leq 2^{|u|-1}\mbox{\@.}
\end{align}
In case $|u|$ is even, there are $2^{|u|-1}$ possible $w$'s such that
$(-1)^{|u| - |w|} = 1$ and also $2^{|u|-1}$ possible $w$'s such that
$(-1)^{|u| - |w|} = -1$\@. The case $|u|$ is odd follows analogously.
% The bound in~(\ref{bound}) is rather coarse and can easily be improved by a more detailed analysis of the maximum number of $1$ and $-1$ coefficients for the summands above.
The main point here is that we obtain an upper bound~(\ref{bound}) which only depends on $u$\@.
Therefore, plugging the bound~(\ref{bound}) into the definition~(\ref{def:shapley.owen}) we  obtain an upper bound for Shapley-Owen values with respect to $\Value^\alpha$\@.
%\begin{proposition}
For $\alpha \neq 0$ and $\Value^{\alpha}$ as defined 
in~(\ref{def:value.alpha})\@:
\begin{align}\label{prop:bounded}
|\Sh{u}(\Value^{\alpha})| 
&\leq \frac{2^{|u| - 1}}{(d - |u| + 1)} 
     \sum_{v \,:\,v \subseteq \Compl{u}} 
        \binom{d - |u|}{|v|}^{-1}  \Def \kappa_u\mbox{\@.}
\end{align}
%\end{proposition}
% Calculating inverses of binomial coefficients
% https://math.stackexchange.com/questions/151441/calculate-sums-of-inverses-of-binomial-coefficients
% Also: Zeilberger's algorithm
This sum of inverse binomial coefficient can also be expressed analytically (e.g. Zeilberger's algorithm)\@.
% {\bf TO DO:} \emph{analytic expression for sum above 
% (Zeilberger's algorithm?).}
For the linearity of Shapley-Owen values and the identity~(\ref{def:sobol}) we obtain a spectral decomposition of Shapley-Owen values from the inequality~(\ref{prop:bounded}) and Lemma~\ref{lem:invariant}\@.
% \begin{align}
% = \Sh{u}(\Value^U) 
%= \sum_{u \subseteq v} \Sh{u}(\sigma^2_v)
%= \sum_{v \subseteq u} \Sh{u}(\sum_{\alpha=0}^\infty y^2_\alpha\,
%\chi_v(\alpha))
%\sum_{v \subseteq u} (\sum_{\alpha=0}^\infty y^2_\alpha\,\Sh{u}
%(\chi_v(\alpha))\mbox{\@.}   
%\end{align}
\begin{theorem}\label{thm:shapley.char}
For all $u \in [1,d]$\@:
\begin{align}\label{char:shapley}
 \Sh{u}(\Value^X) &=
    \sum_{\alpha=1}^\infty y^2_\alpha\,\Sh{u}(\Value^{\alpha})\mbox{\@.}
\end{align}
\end{theorem}
\noindent
The infinite sum on the right-hand side is well-defined, since
$\sum_{\alpha=1}^\infty y_{\alpha}^2$ converges 
and $|\Sh{u}(\Value^{\alpha})|$ is bounded 
by~(\ref{prop:bounded})\@.
The significance of the identity~(\ref{char:shapley})
is that the elementary Shapley values $\Sh{u}(\Value^\alpha)$
in the identity~(\ref{char:shapley})
depend only on the normalized Legendre polynomials $\widetilde{\Le{\alpha}}(X)$ 
but not on $\M(X)$\@,
and can therefore be precomputed once and for all.
In addition, as long as $\M(X)$ and the underlying distribution of $X$ are unchanged, the PCE of $\M(X)$
can be reused in a fairness analysis of $\M(X)$ to compute other Shapley-Owen effects.
The price we pay is the outer infinite sum in~(\ref{char:shapley}), but we can at least approximate this infinite sum using a truncated PCE of $(\M \circ \T^{-1})(U)$ with $l$ summands\@:
  \begin{align}\label{approx:shapley}
 \widehat{\Sh{u}}^{(l)}(\Value^X) &=
    \sum_{\alpha=1}^l y^2_\alpha\,\Sh{u}(\Value^{\alpha})\mbox{\@.}
\end{align}
Therefore, from Theorem~\ref{thm:shapley.char} and the identities in Section~(\ref{subsec:truncated.PCE}) on the truncation error, we obtain an upper bound on the approximation error of (\ref{approx:shapley})\@.
\begin{corollary}\label{prop:bound.shapley}
For $\epsilon_l \Def \sum_{\alpha = l+1}^\infty y_{\alpha}^2$ as defined 
in~(\ref{subsec:truncated.PCE}) and the upper bound $\kappa_u$ as defined in Proposition~\ref{prop:bounded}\@:
\begin{align}
    | \Sh{u}(\Value^X) -  \widehat{\Sh{u}}^{(l)}(\Value^X)|
&\leq \kappa_u \epsilon_l\mbox{\@,}
\end{align}
and, a fortiori,  
$\lim_{l \to \infty}| \Sh{u}(\Value^X) -  \widehat{\Sh{u}}^{(l)}(\Value^X)| = 0$\@.
\end{corollary}

We assume the elementary 
Shapley-Owen values $\Sh{u}(\chi_v(\alpha)) = 
\Sh{u}(\chi_v(\Le{\alpha}(X)))$ to be tabulated for all
$u \subseteq [1,d]$ and sufficiently many (see below) multi-indices $\alpha$\@.
Now, the Shapley-Owen effects for $\M(X)$ therefore are computed as follows:
\begin{enumerate}
    \item Compute a  truncated and sparse PCE
    $\sum_{\alpha = 0}^l y_\alpha \widehat{\Le{\alpha}}(U)$, for $l \in \Nat$, 
    of $\mathcal{M} \circ \T^{-1}(U)$ which is precise up to some given $\epsilon > 0$ (Lemma~\ref{prop:stoppage.criterion} using the algorithm suggested in Section~(\ref{subsec:sparse.PCE})\@.
    % The error for approximating the variance of the partial effects of $\mathcal{M} \circ \T^{-1}(U)$
    % by $\hat{\sigma}_v^2$ is also bounded by $\epsilon$
    % (Lemma~(\ref{approximated.variance.total.effects}))\@.
    \item Compute the approximated Shapley-Owen effect 
            $\widehat{\Sh{u}}^{(l)}(\Value^X)$~(\ref{approx:shapley})\@.
    
   % $\hat{\Sh{u}}(\Value^X)$ by the truncated summation
   %    \begin{align}
   %    \sum_{v:\,v \subseteq u} (\sum_{\alpha=0}^l y^2_\alpha\,\Sh{u}(\chi_v(\alpha))\mbox{\@.}
   %   \end{align}
\end{enumerate}
The total error of the approximated Shapley-Owen effect in Step~(2) is bound by $\epsilon\kappa_u$ (Corollary~\ref{prop:bound.shapley})\@.
Note that the coefficients $\kappa_u$ are exponential in the 
cardinality $|u|$, but $|u|$ is typically rather small in fairness analysis. For example, $|u|$ is between $2$ and $5$ for the popular fairness datasets surveyed in~\cite{gohar2023survey}\@.
The main computational challenge in this algorithm is the construction of a sufficiently precise but sparse truncated PCE for $(M \circ \T^{-1})(U)$ in step (1)\@.
% A number of solutions have been developed in the field of \emph{uncertainty quantification}\@.
% The approximate model is usually validated by an 
% \emph{a posteriori} error estimation.
% In contrast, we suggest in Section~\ref{subsec:truncated.PCE} an iterative algorithm for computing a truncated and sparse PCE until a given error bound $\epsilon$ is satisfied\@.

\newcommand{\SC}{\mathit{SC}}

\section{Related Work}\label{sec:related}

Improving algorithmic fairness in machine learning 
typically works by de-biasing the set of training data, by incentivizing learning to make fair decisions, or by removing bias from the output of the machine learning model~\cite{bellamy2019ai,bird2020fairlearn,dwork2012fairness,hardt2016equality,jagielski2019differentially,kearns2018preventing,konstantinov2022fairness,kusner2017counterfactual,mehrabi2021survey,sharifi2019average,wexler2019if,zemel2013learning}\@. 
These are all interventions in the design of the machine learning model. % , whereas Shapley fairness validates the model used.
A major drawback is that these methods typically embed a fixed notion of fairness into the decision making, whereas the concept of fairness is context-specific.
In contrast, measuring fairness in terms of Shapley-Owen effects allows explicitly specifying fairness constraints and verifying the underlying algorithm with respect to these constraints.

Techniques inspired by formal methods have been used to  establish algorithmic fairness properties
by verifying ~\cite{albarghouthi2017fairsquare,bastani2019probabilistic,ghosh2021justicia,meyer2021certifying,sun2021probabilistic} 
or enforcing~\cite{balunovic2021fair,ghosh2022algorithmic,john2020verifying}
a machine-learned model
on all runs of the system with high probability. 
This requires certain knowledge about the system model, which may not always be available, 
whereas our sensitivity analysis only requires the computability of a square-integrable model under analysis. In contrast to all this work, our fairness analysis treats the model as a black box.
Runtime verification, which is a lightweight solution for checking properties based on a single, possibly long execution trace of a given system, has recently been adapted for the problem of bias detection in decision functionality~\cite{henzinger2023monitoring}\@.
Like our approach, runtime monitoring is largely agnostic to the structure of the model, but it is limited to detecting bias and enforcing fairness from a single run.

% One of the advantages of formulating feature 
% importance measures through Shapley attribution values
% is the flexibility in choosing the value function.
% Sensitivity analysis is a popular technique for analyzing machine-learned models~\cite{borgonovo2024many,olsen2023comparative}\@.
Shapley values are a popular technique for explaining individual 
input contribution in black-box predictive machine-learned models~\cite{aas2021explaining,datta2016algorithmic,lundberg2020local,lundberg2017unified,molnar2020interpretable,vstrumbelj2014explaining, sundararajan2020many,olsen2023comparative,borgonovo2024many}\@.
% The \emph{baseline value} function, for example quantifies the difference of 
% the prediction with respect to a fixed base case when certain subsets of features are modified~\cite{sundararajan2020many}\@. 
% The corresponding Shapley attribution is \emph{local} in that it focuses on a single observation,
% and it also does not take feature uncertainty into consideration.
These Shapley-based analyses are usually \emph{local} in that they are based on the conditional expectation $\CE{\M(X)}{X_u = x_u^*}$ 
for a single observation $x_u^*$\@.
% The conditional expectation is the minimizer of the commonly used squared error loss function.
% and it also does not take feature uncertainty into consideration.
Corresponding global values % $\E{\CE{\M(X)}{X_u}}$ 
also consider input uncertainty 
by averaging the local value function over initial points~\cite{strumbelj2010efficient,vstrumbelj2014explaining}\@.
Most importantly, the so-called SHAP values~\cite{lundberg2017unified}, 
which are widely used as \emph{post hoc} explanations for the results of machine learning models,
are based on individual Shapley attributions for conditional expectation values.
% Monte Carlo integration and regression are typically used to 
% compute conditional expectations.
Assuming that the inputs are independent simplifies computations such as Monte Carlo integration and regression~\cite{lundberg2017unified,aas2021explaining}\@,
%However, this independence assumption may not be generally true and 
but may also lead to
incorrect explanations~\cite{aas2021explaining,merrick2020explanation,fryer2021shapley,olsen2022using}\@.
Such misinterpretations are unacceptable for a sensitive topic like fairness, and we therefore do not make simplifying assumptions such as input independence. In addition, the underlying fairness measures are computed up to known error bounds.

Unlike SHAP, our developments rely on Shapley-Owen effects, which use the variance-based value $\V{\CE{\M(X)}{X_u}}$ to assign relative importance to subsets of inputs.
This variance-based value function $\Sobol{u}$ for quantifying the relative importance has originally been suggested by~\cite{owen2014sobol}\@.
For linear model with independent inputs the 
Shapley regression value $R^2_u$~\cite{lipovetsky2001analysis,gromping2007estimators,huettner2012axiomatic,gromping2015variable} for feature importance equals
the normalized Sobol' index $\nicefrac{\Sobol{u}}{\sigma^2}$~\cite{saltelli2000sensitivity}\@.
This variance-based value has been applied to the sensitivity analysis of neural networks~\cite{fock2013global,fernandez2016global,kowalski2017sensitivity,kowalski2018determining,li2018first,cheng2019neural}\@. 
The two-features Shapley-Owen value
% in Example~\ref{ex:two-shapley-owen} 
has been applied for the explanation of tree-based machine learning models~\cite{lundberg2020local}\@.
The variance-based value allows one to define variance-based sensitivity measures for models with 
dependent inputs~\cite{song2016shapley,owen2017shapley,benoumechiara2019shapley,iooss2019shapley}\@.

The defining identity~(\ref{def:shapley.owen}) of Shapley-Owen effects $\Sh{u}(\Value)$ requires a summation over all supersets $v$ of the complement 
of $u$ together with the computation of 
the relative importance $\V{\CE{\M(X)}{X_v}}$\@. 
This exponential number of summands in identity~(\ref{eq:shapley-owen}) severely limits applicability\@.
Alternatively, estimation algorithms, based on consistent estimators of the Shapley effects, have been proposed~\cite{song2016shapley}\@.
The \emph{exact permutation} algorithm  is expensive, since it traverses the exponential space of permutations between the inputs,
whereas the \emph{random permutation} algorithm is approximative in that it is based of randomly sampling some permutations of the inputs.
Experimental results with these estimators for Shapley effects are reported in~\cite{mara2021polynomial}\@.
The characterization of the Shapley-Owen value in terms of its M\"obius inverse (Appendix~\ref{app:moebius}) still requires an exponential number of summands~\cite{plischke2021computing}\@. 
Our algorithm is novel in that it uses a spectral decomposition of Shapley-Owen effects to separate the computation into model-specific and model-independent computations. In this way, the computation of a Shapley-Owen effect is essentially reduced to the PCE of the underlying model, which is one of the central problems in \emph{uncertainty quantification}\@.
Our algorithm is also unique in that it provides an upper bound on the accumulated errors in the computation of Shapley-Owen effects.
These error bounds are especially important 
for the sensitive issue of fairness analysis.

\section{Conclusions}\label{sec:conclude}

We argue that the fairness of algorithmic decisions needs to be evaluated according to different dimensions of the inputs and how they interact.
If we accept a small number of conditions for equitable attributions, then Shapley-Owen effects are the only way to measure fairness.
There is no assumption of input independence in the calculation of Shapley-Owen effects, as this type of unrealistic assumption leads to incorrect fairness conclusions.
% To make the computation of Shapley-Owen effects tractable 
% we make the basic assumption that higher order interactions are negligible. 
% Of course, such an assumption must be validated on a case-by-case basis.

Variance-based measures for assigning relative importance to inputs form the basis of \emph{global fairness analysis},
which
% , similar to \emph{global sensitivity analysis}, 
uses variance decomposition to quantify which inputs and their interactions most influence unfair decisions.
In this way, for a given set of fairness constraints on sensitive inputs, an algorithmic decision can be validated to be fair with respect to those inputs. 
If the Shapley-Owen allocation of the relative importance of some sensitive inputs is small enough to be considered fair, those inputs can be projected away or, alternatively, set to a fixed value.
Fairness analysis based on Shapley-Owen effects can also detect interactions among sensitive inputs that would not be detected by simply varying the inputs individually.

\begin{appendix}

\section{ANOVA Decomposition}\label{app:anova}
Let $X = (X_1, \ldots,, X_d)$ be a $d$-dimensional real-valued vector of independent random variable.
The pdf of $X$ is denoted by $f$\@.
$\Psi_{\alpha}(X)$, for $\alpha$ an enumerable, infinite set, is an orthogonal basis of
$L^2(\Omega)$ with respect to the inner product
$\Inner{.}{.}{f}$ of weight $f$\@.
Then the square-integrable $\M(X)$ can be decomposed 
as
\begin{align}
    \M(X) &= \sum_{u \subseteq [1,d]} \M_u(X_u)\mbox{\@,}
\end{align}
where the independent effects $\M_u(X_u)$ only depend on the subset $X_u$ of inputs.
They are defined recursively by 
\begin{align*}
\M_{\emptyset}(x_{\emptyset}) &= \E{\M(X)} \\
\M_{u}(x_{u}) 
&= \int_{\Real^{|\Compl{u}|}}
         (\M(x) - \sum_{v \subsetneq u} \M_v(x_v))\,f_X(x)
        \,d x_{\Compl{u}} \\
&= \int_{\Real^{|\Compl{u}|}} \M(x)\,f_X(x) \,d x_{\Compl{u}}
    - \sum_{v \subsetneq u} \M_v(x_v))\mbox{\@.}
%&= \CE{\M(X)}{X_{\Compl{u}}} - %\sum_{v \subsetneq u} \M_v(x_v))
\end{align*}

\section{Marginal Cdfs}\label{app:marginal.cdfs}

The marginal cdfs 
   $F_k(x_k \,|\,x_{[1,k-1]}) \Def \Prob{X_k \leq x_k \,|\, X_1 = x_1, \ldots, X_{k-1} = x_k}$
can be obtained analytically from the pdf $f_X(x)$ of $X$\@.
First, the marginal pdf $f_{X_k}(.)$ is given by
\begin{align}
    f_{X_k}(x_{[1,k]}) 
    &= \int_{\Real^{d - k +1}}
          f_X(x) \,dx_{[k+1,d]}\mbox{\@.}
\end{align}
Second, the conditional pdf $f_k(.)$ is defined in terms of the marginal pdfs:
\begin{align}
    f_k(x_k \,|\, x_{[1,k-1]})
  &= \frac{f_{x_k}(x_{[1,k]}) }
          {f_{x_{k-1}}(x_{[1,k-1]})}\mbox{\@.}
\end{align}
Third, the marginal cdfs $F_k(.)$ are obtained by integrating the conditional pdf $f_k(.)$ over $x_k$\@:
\begin{align}
    F_k(x_k \,|\, x_{[1,k-1]})
  &= \frac{\int_{-\infty}^{\infty} f_{x_k}(x_{[1,k-1]}, t) \,dt }
          {f_{x_{k-1}}(x_{[1,k-1]})}\mbox{\@.}
\end{align}

\section{Model Validation}\label{app:model.validation}

In practice, an \emph{a posteriori} error estimate of $\widehat{\M}(X)$, when the inputs $X$ are dependent,
is often calculated 
with the \emph{leave-one-out} cross-validation
method from model diagnosis.
Let $\Sample{x}{k}$, for $k = 1, \ldots, N$, be 
an experiment, and $\widehat{\M}_{\sim k}$ be an approximation of the PCE for the truncated $\widehat{\M}$ which is obtained from all samples except for $k$ (say, by regression)\@.
The corresponding \emph{leave-one-out}
error is then estimated by the mean predicted residual sum of squares.
\begin{align}
    \mathit{l1o} &\Def \nicefrac{1}{N} \sum_{k=1}^N (\M(\Sample{x}{k}) - \widehat{\M}_{\Compl{k}}(\Sample{x}{k}))^2\mbox{\@.}
\end{align}
The larger the value of $1 - \nicefrac{\mathit{l1o}}{\sigma^2}$
the more accurate the constructed PCE approximation.

\section{Shapley-Owen Effects}\label{app:so.effects}

We list some special cases for which Shapley effects are easy to compute.

\begin{example}
   If $\M(X) = a + b^T X$ for $a \in \Real$, $b = (b_1, \ldots, b_d) \in \Real^d$, and variables $X_i$, for $i = 1,\ldots,d$, independent with variance $\V{X_i} < \infty$
        then $\Sh{i} = b_i^2 \V{X_i}$\@.
       Similarly, if $\M(X) = \sum_{i=1}^d \N_i(X_i)$ and independent variables $X_i$, for $i = 1,\ldots,d$
        then $\Sh{i} = \V{\N_i(X_i)}$\@.
\end{example}

\begin{example}
If $\V{\CE{\M(X)}{X_u}} = 0$ for $|u| > 2$ then
the  game is represented by the induced subgraph with nodes $1, \ldots, d$ and weights
$(w_{ji} =)~w_{ij} \Def \V{\CE{\M(X)}{X_{ij}}}$, for $1 \leq i, j \leq d $ on the edges between nodes $i$ and $j$\@.
For the game $\Val{u} \Def \sum_{1 \leq i, j \leq d} w_{ij}$, the Shapley value $\Sh{i}(\Value)$ equals~\cite{deng1994complexity}
   \begin{align}
       \Sh{i}(\Value) &= w_{ii} + \sum_{j \in \Compl{i}} w_{ij}\mbox{\@.}
   \end{align}
Hence, it can be computed in polynomial time. 
This is easy to see, since by additivity~(\ref{Sh.additivity}) 
one can consider each edge separately. 
Assume the single edge $E \Def \{\{i,j\}\}$\@. 
Clearly, $\Compl{\{\{i,j\}}$ are dummy players, so from 
the dummy~(\ref{Sh.null}) and the efficiency~(\ref{Sh.efficiency})
properties of Shapley values we get
$\Sh{i}(val) = \Sh{j}(val) = \nicefrac{1}{2} w_{ij}$\@.
If $i = j$ then, analogously, $\Sh{i}(val) = w_{ii}$\@.
\end{example}

\begin{example}[\cite{owen2017shapley}, Theorem 2]
If $\M(X) = a + b^T X$ for 
$a \in Real$, $b = (b_1, \ldots, b_d) \in \Real^d$, 
and $X \sim \mathcal{N}(\mu, \Sigma)$, where $\Sigma^{d \times d}$ has full rank, then
\begin{align}
      \Sh{i} &= \frac{1}{d}
               \sum_{u \,:\, u \subseteq \Compl{i}} {\binom{d - 1}{|u|}}^{-1} 
               \frac{{\Cov{X_j}{b^T_{\Compl{u}} X_{\Compl{u}} \,|\,X_u }}^2}
                    {\CV{X_j}{X_u}}\mbox{\@.}
\end{align}
\end{example}

\section{PCE for Dependent Inputs}\label{app:dependent.PCE}

For  marginal pdfs $f_i$ of the input variables $X_i$, for $i = 1, \ldots, d$\@,
the multi-index $\alpha = (\alpha_1, \ldots, \alpha_d)$,
and
univariate orthogonal polynomials $P_{\alpha_i}(x_i) \in \Real[x_i]$ of degree $\alpha_i \in \Nat$, for $i = 1, \ldots, d$\@,
the chaos function $\Psi_{\alpha}(X)$ is defined to be
\begin{align}\label{pce:dependent.basis}
  \Psi_{\alpha}(x) &\Def  
     \sqrt{\frac{ \prod_{i=1}^d f_i(x_i)}{f(x)}}%^{\nicefrac{1}{2}} \,
     \prod_{i=1}^d P_{\alpha_i}(x_i)\mbox{\@,}
\end{align}
as long as $f(x) \neq 0$\@.
For independent inputs, the definition~(\ref{pce:dependent.basis}) reduces to the definition~(\ref{multivariate.independent})\@.
The family $(\Psi_{\alpha})_{\alpha \in \Nat^d}$ as defined in~(\ref{pce:dependent.basis}) is a complete orthonormal basis for $L^2(\Omega)$~(\cite{soize2004physical};\cite{sullivan2015introduction}, Chapter 11.3)\@.
Note that the $\Psi_{\alpha}$, for $\alpha \in \Nat^d \setminus \{0\}$, are not polynomial.
But we can still have a chaos expansion as in~(\ref{def:pce}), where the expectation and the variance of $\M(X)$ is obtained analytically from the expansion coefficients as in (\ref{pce:expectation}) and (\ref{pce:variance})\@. 
Alternatively, a complete orthogonal polynomial basis of $L^2(\Omega)$ is constructed by Gram-Schmid orthogonalization from the complete monomial basis $(X^{\alpha})_{\alpha\in\Nat^d}$ of $L^2(\Omega)$~ (Appendix~\ref{app:gram-schmid})\@.

\section{Gram-Schmid Orthogonalization}\label{app:gram-schmid}

Applying Gram-Schmid orthogonalization to
the enumerable set $\{X^\alpha \,|\, \alpha \in \Nat^d\}$ of monomials
yields a complete orthogonal polynomial basis of $L^2(\Real^d)$\@.
\begin{align}
    \Psi_0(X) = 1 &,~~~~~~~~~
    \Psi_i(X) =
       m_i(X) - \sum_{k=0}^{i-1} c_{ik} \Psi_k(X)\mbox{\@,}
\end{align}
for $i\leq 1$, $m_i(X)$ the $i$th monomial of the form $X^\alpha$, for some multi-index $\alpha \in \Nat^d$, in some fixed enumeration, and the projection coefficients
  \begin{align}\label{def:proj}
  c_{ik} &= \frac{\Inner{m_i(X)}{\Psi_k(X)}{f}}{\Norm{\Psi_k(X)}{f}}
  \end{align} 
Calculating the inner products
~(\ref{def:proj}) typically hinges on the functional form of the pdf $f$ of the input $X$\@. 
But these coefficients can alternatively also be computed 
via direct substitution of the moments of $X$ as follows\@.
For a multi-index $\alpha \in \Nat^d$ the
\emph{multi-variate  moment}
\begin{align}
   \mu^\alpha &\Def \int_0^\infty X^\alpha f(X)\, dX\mbox{\@.} 
\end{align}
is easily determined using either sampling 
or the functional form of the pdf $f$ of $X$\@.
% From the definitions it is also easy to see that
Since
$\Inner{X^\alpha}{X^\beta}{f} = \mu^{\alpha + \beta}$,
% Moreover,  $\Psi_i(X)$ is, by construction, a  linear combination of monomials $m_k(X)$ for $k \leq i$\@.
the projection coefficients $c_{ik}$ are 
computed recursively from the multi-variate moments, without further reference of the pdf $f$~\cite{mesbahpaulson2017} of the inputs $X$\@.
% For the construction of multivariate orthonormal polynomials from moments see also~\cite{golub1969calculation}\@.
% Note that all the inner product computations may
% rely on the raw moments of $X$ only. 
  
% Does this aX_1^\alpha_1lso work for multivariate?
% For a proof see, for example: 
% https://math.stackexchange.com/questions/1598343/polynomials-form-as-hilbert-basis-for-l2
% An orthogonal basis of the $L^2(\Omega, \Real)$ vector space
% in terms of (univariate!) monic polynomials $\pi_i(x) \in \Real[x]$, for $i \in \Nat$,
% is constructed by the \emph{3-term recurrence 
% relation}
% \begin{align}
%  \pi_{i+1}(x) &= (x - a_k) \pi_k(x) - b_k \pi_{k-1}(x)
%\end{align}
%with $\pi_{-1}(x) = 0$, $\pi_{0}(x) = 1$, and the coefficients $a_k$, $b_k$ are determined by
%\begin{align}
%    a_k &= \nicefrac{\Inner{x\pi_k}{\pi_k}{f}}{\Inner{\pi_k}{\pi_k}{f}} \\
%    b_k &= \nicefrac{\Inner{\pi_k}{\pi_k}{f}}{\Inner{\pi_{k-1}}{\pi_{k-1}}{f}}\mbox{\@.}
%\end{align}
%Gram-Schmidt orthogonalization satisfies this recurrence.

\section{Computing Galerkin Projection}\label{app:galerkin}

Suppose that the random variable $X$ ($\M(X)$) has a continuous distribution for which the cdf is $F$ ($G$)\@.
Then, using the \emph{probability integral transform},
$U = F(X) = G(\M(X))$ for $U$ a uniformly
distributed random variable supported on $[0,1]^d$\@.
% are isoprobabilistic transforms.
The Galerkin projection~(\ref{Galerkin}) is
now expressed as
\begin{align}
   y_\alpha 
    = \E{\M(X)\, \Psi_{\alpha}(X)}
   = \int_{[0,1]^d} G^{-1}(u) \Psi_{\alpha}(F^{-1}(u)) du\mbox{\@,}
\end{align}
where $F^{-1}$ and $G^{-1}$ are the generalized inverses of the isoprobabilistic transforms $F$ and $G$, respectively\@.

\section{Truncation Error}\label{app:truncation.error}

Let $\M(X) = \sum_{\alpha=0}^\infty y_\alpha \Psi_{\alpha}(X)$ be a PCE 
based on the orthonormal sequence $(\Psi_{\alpha}(X))_{\alpha \in \Nat}$, 
$\widehat{M}_l(X)$ is a truncation of the PCE $\M(X)$ with $l+1$ terms\@,
and $\epsilon_l \Def \M(X) - \widehat{\M}_l(X)$ the corresponding truncation error; then:
\begin{align*}
 & \E{(\M(X) - \widehat{\M}_l(X))\,\widehat{\M}_l(X)} \\
=~ & \E{(\sum_{\alpha=l+1}^\infty y_\alpha \Psi_{\alpha}(X))\,
        \sum_{\beta=0}^l y_\beta \Psi_{\beta}(X)} \\
=~ & \E{\sum_{l+1 \leq \alpha < \infty}\sum_{0 \leq \beta \leq l}
               y_\alpha y_\beta \Psi_{\alpha}(X) \Psi_{\beta}(X) } \\
=~ & \sum_{l+1 \leq \alpha < \infty}\sum_{0 \leq \beta \leq l}
               y_\alpha y_\beta \E{\Psi_{\alpha}(X) \Psi_{\beta}(X)} \\
=~ & 0\mbox{\@.}
\end{align*}
Therefore the truncation error $\epsilon_l$ is orthogonal to the truncated PCE\@.

\section{Examples of Cooperation Games}\label{app:cooperation.games}

% Let $\M$ be a function of $d \in \Nat$ independent random variables given by $X = (X_1, \ldots, X_d)$\@, 
%and let $\Val{u} \in \Real$ be
%the value attained by the subset 
%$u \subseteq \{1, \ldots, d\} \equiv D$\@.
%It is always assumed 
%that $\Val{\emptyset} = 0$, and we use the notation $\Compl{u}$ for the set difference $D \setminus u$\@. 
%The \emph{Shapley} value $\Sh{i}(\mathit{val})$ for 
%$i = 1, \ldots, d$ is
%  \begin{align} \label{def:shapley}
%      \Sh{i}(\Value) &= \nicefrac{1}{d}
%               \sum_{u \,:\, u \subseteq \Compl{i}} {\binom{d - 1}{|u|}}^{-1} (\Val{u + i} - \Val{u})\mbox{\@,}
%  \end{align}
%where $u + i$ is shorthand for $u \cup \{i\}$\@. 
%The \emph{marginal contribution} $(\Val{u + i} - \Val{u})$ is how much the model changes when a new feature $i$ is added,
%the \emph{combinatorial weight} ${\binom{d - 1}{|u|}}$$^{-1}$ is the weight given to each of the different subsets of features with size $|u|$\@,
%and \emph{averaging} by $\nicefrac{1}{d}$ determines the average of all marginal contributions from all possible subsets of sizes ranging from $0$ to $d-1$\@.

\begin{example}\label{ex:simple}
Consider the 3-player game with coalition values:\\
 \begin{center}
\begin{tabular}{|c||c|c|c|c|c|c|c|c|}\hline
          & $\emptyset$ & $\{1\}$ & $\{2\}$ & $\{3\}$  & $\{1,2\}$ & $\{1,3\}$ & $\{2,3\}$ &  $\{1, 2,3\}$ \\\hline
  $\Val{.}$ & $0$ & $0$ & $2$ & $0$ & $5$ & $6$ & $7$ & $10$ \\\hline
\end{tabular}
\end{center}
The Shapley attributions are
  \begin{align*}
   \Sh{1}(\Value) &= 
     \nicefrac{1}{3} \cdot
     ( 1 \cdot (0 - 0) + 
      \nicefrac{1}{2} \cdot (5 - 2) + 
       \nicefrac{1}{2} \cdot (6 - 0) + 
       1 \cdot (10 - 7)
     )
   = \nicefrac{5}{2}\\
   \Sh{2}(\Value) &=
   \nicefrac{1}{3} \cdot
   (1 \cdot (2 - 0) + 
    \nicefrac{1}{2} \cdot (5 - 0) + 
    \nicefrac{1}{2} \cdot (7 - 0) +
    1 \cdot (10 - 6)
   )
   =   4\\
   \Sh{3}(\Value) &=
    \nicefrac{1}{3} \cdot
   (1 \cdot (0 - 0) + 
    \nicefrac{1}{2} \cdot (6 - 0) + 
    \nicefrac{1}{2} \cdot (7 - 2) +
    1 \cdot (10 - 5)
   )
   = \nicefrac{7}{2}\mbox{\@.}
   \end{align*}
\end{example}
\noindent
Alternatively, the Shapley attribution is characterized by
  \begin{align}\label{def:shapley.permutation}
  \Sh{i}(\Value) &=
    \nicefrac{1}{d!} \sum_{\sigma \in \Pi(D)}
        \Val{\Restrict{\sigma}{i} + \,\sigma(i)} - \Val{\Restrict{\sigma}{i}}\mbox{\@,}
  \end{align}
where $\Pi(D)$ is the set of permutations of elements in $D$,
$\Restrict{\sigma}{1} \Def \emptyset$, and, for $i > 1$,
the prefix
$\Restrict{\sigma}{i} \Def \{\sigma(1), \ldots, \sigma(k - 1 )\}$ for $k$ such that $\sigma(k) = i$\@.

In economics, the Shapley value is widely used to solve the attribution problem, when the value produced through the joint efforts of a team is to be attributed to individual members of that team.

A cooperation game with value $\Value$ can be written in a unique way as a linear combination 
of the form %~\cite{van2022approximation}%\footnote{
%Attributed to: Borm, \emph{Games, cooperative behavior, 
%and economics}, 2010.
%}
   \begin{align}
   \Val{u} &= \sum_{v\,:\,v \subseteq D \setminus\emptyset} c_v \UG_v(u)\mbox{\@,}
   \end{align}
where $c_v \in \Real$ and $U_v$ is the \emph{unanimaty game}
    \begin{align}
    \UG_v(u) &\Def \begin{cases}
        1 & \text{if~} v \subseteq u \\
        0 & \text{otherwise}\mbox{\@.}
    \end{cases}
    \end{align}
It can also be shown that the constants $c_v$ coincide with \emph{Harsanyi dividens}\@.
Using the additivity property the Shapley attribution of every cooperation game can therefore be obtained from the Shapley attributions of unanimity games~\cite{shapley1953value}\@:
   \begin{align}\label{eq:shapley.decompose}
       \Sh{i}(\Value) &= 
           \sum_{v\,:\,i \in v}
               \frac{c_v}{|v|}\mbox{\@.}
   \end{align}
\begin{example}
The game in Example~\ref{ex:simple} can be described as 
the linear combination of unanimity games:
\newcommand{\ug}[1]{\UG_{\{#1\}}}
   $\Val{u} = 2\ug{2}(u)+3\ug{1,2}(u)+6\ug{1,3}(u)+5\ug{2,3}(u)-6\ug{1,2,3}(u)
   $\@.
The three Shapley values, in vectorial notation, are obtained from equation~(\ref{eq:shapley.decompose})\@.
\newcommand{\Vecthree}[3]{\begin{pmatrix} {#1} \\ {#2} \\ {#3} \end{pmatrix}}
\begin{align*}
\Vecthree{\Sh{1}}{\Sh{2}}{\Sh{3}}
&= 2  \Vecthree{0}{1}{0}  + 
  \nicefrac{3}{2}  \Vecthree{1}{1}{0} + 
  3 \Vecthree{1}{0}{1} +
  \nicefrac{5}{2}  \Vecthree{0}{1}{1} + 
  2 \Vecthree{0}{1}{1}  ~=~
  \Vecthree{\nicefrac{5}{2}}{4}{\nicefrac{7}{2}} 
\end{align*}
\end{example}

\section{M\"obius Inverse of Shapley-Owen Effects}\label{app:moebius}

Using the variance-based value function $\Val{u} \Def \V{\CE{\M(X)}{X_u}}$ and assuming 
input independence one obtains the characterization
\begin{align}\label{eq:shapley}
    \Sh{i} &= \sum_{v \,:\,i \in v} \frac{\sigma^2_v}{|v|}
\end{align}
from the Shapley properties~\cite{owen2014sobol} or, alternatively,
using the M\"obius inverse of the Sobol' indeces $S_u$\@.
In the framework of independent input variables, it means that the Shapley 
effect $\Sh{i}$ associated to input $X_i$ shares the part of variance due 
to the interaction of a subset $u \subseteq D$, for $i \in u$, of inputs $X_u$, 
with each individual input within that subset. 
% The values $\Sobol{u}$ and $\SobolT{u}$ lead to identical Shapley 
% effects~\cite{song2016shapley}\@.
Now, plugging the identity~(\ref{variance.total.effects})
into the characterization~(\ref{eq:shapley}) one can express Shapley values in terms of the coefficients of the PCE of $\M$\@.

% Total Sobol indeces satisfy the equality
%   \begin{align}
%      \SobolT{i} &= \sum_{i \in v} \Var{Y_i}\mbox{\@.}
%   \end{align}
% Now, % since  $S_i = \Var{Y_i}$
From the characterization~(\ref{eq:shapley})
of Shapley-Owen effects in the presence of input independence
one obtains \emph{bracketing} inequalities $\Sobol{i} \leq \Sh{i} \leq \SobolT{i}$ for the Shapley attribution in terms of the correponding Sobol' indices as defined in~(\ref{def:sobol}) and~(\ref{def:sobolT})\@.
%EThe upper bound of this inequality can be sharpened to 
%$\Sh{i} \leq \nicefrac{1}{2} (\Sobol{i} + \SobolT{i})$\@.
So the need for computing Shapley effects arises only if the gap between 
first order and total effects is large or dependencies in the input are present. 
But these bounds do
not hold anymore in case the input factors present some dependencies~\cite{song2016shapley,iooss2019shapley}\@.
%For linear Gaussian models of dimension $2$, for example, the Shapley effects are equal to the first-order and total Sobol’ indices~(\cite{iooss2019shapley}, Section 3.2)\@.
In the presence of correlation, the Shapley effects for a linear Gaussian model of dimension $2$, 
for example,  lie between
the full first-order indices and the independent total indices: with either 
$\Sobol{i} \leq \Sh{i} \leq \SobolT{i}$
or
$\SobolT{i} \leq \Sh{i} \leq \Sobol{i}$~(\cite{iooss2019shapley}, Section 3.2)\@.

Analogously to the Shapley value in equation~(\ref{eq:shapley}), 
the Shapley-Owen effect can also be expressed in terms of the M\"obius 
inverse of the chosen value function.
Therefore, for the variance-based value function 
   $\Val{u} \Def \V{\CE{\M(X)}{X_u}}$
the Shapley-Owen index~(\ref{def:shapley.owen})
of the inputs $X_u$ can be written as (\cite{grabisch1999axiomatic,rabitti2019})
\begin{align}\label{eq:shapley-owen}
    \Sh{u} &= \sum_{v \,:\,v \supseteq u} \frac{\sigma_v^2}{|v| - |u| + 1}\mbox{\@.}
\end{align}
Using the identity~(\ref{variance.total.effects}), 
Shapley-Owen effects can now be obtained, at any order, from the coefficients of the PCE expansion of $\M$\@.
Identity~(\ref{eq:shapley-owen}) leads to a bracketing inequality 
  %\begin{align}\label{shapley.owen.bracket}
    $\sigma^2_u \leq \Sh{u} \leq \gamma_u$\@,
 % \end{align}
for Shapley-Owen effects, 
where $\gamma_u \Def \sum_{v \,:\,v \supseteq u} \sigma^2_v$ is the 
\emph{superset importance measure} of~\cite{liu2006estimating}\@.
It quantifies the global impact of the input subgroup $X_u$ in all higher-order
Sobol' interaction terms.
The upper bound in this inequality can be sharpened to 
$\Sh{u} \leq  \nicefrac{1}{2}(\Sobol{u} + \gamma_u)$~\cite{plischke2021computing}\@.

\end{appendix}

\bibliographystyle{ACM-Reference-Format}
\bibliography{main}

\end{document}